\documentclass[a4paper,fleqn]{cas-sc}

\usepackage[authoryear]{natbib}

\usepackage{booktabs} 
\usepackage{multirow}
\usepackage{graphicx}
\usepackage{subcaption}
\usepackage{url}

\usepackage{tabularx}
\usepackage{array}

\usepackage{enumitem}

\usepackage{hyperref}

\hypersetup{
    colorlinks=true,
    linkcolor=blue,
    citecolor=blue,
    urlcolor=blue
}

\begin{document}
\let\WriteBookmarks\relax

\shorttitle{From Affect Prediction to Affect Forecasting}
\shortauthors{S. Noor et~al.}

\title[mode=title]{From Affect Prediction to Affect Forecasting: Evidence for Distinct Information Sources in Longitudinal Text}

\author[1]{Sadia Noor}
\credit{Conceptualization, Methodology, Software, Formal Analysis, Investigation, Data Curation, Visualization, Writing -- Original Draft}

\author[1]{Seemab Latif}
\ead{seemab.latif@seecs.edu.pk}
\credit{Methodology, Validation, Writing -- Review \& Editing}

\author[2]{Raja Khurram Shahzad}
\cormark[1]
\ead{raja-khurram.shahzad@miun.se}
\credit{Methodology, Validation, Writing -- Review \& Editing}

\author[1]{Mehwish Fatima}
\cormark[1]
\ead{mehwish.fatima@seecs.edu.pk}
\credit{Conceptualization, Supervision, Methodology, Validation, Writing -- Review \& Editing}

\affiliation[1]{
organization={School of Electrical Engineering and Computer Science (SEECS)},
addressline={National University of Sciences and Technology (NUST)},
city={Islamabad},
country={Pakistan}
}

\affiliation[2]{
organization={Department of Communication, Quality Management and Information Systems},
addressline={Mid Sweden University},
city={Ostersund},
country={Sweden}
}

\cortext[cor1]{Corresponding authors}


\begin{abstract}
Modeling dimensional affect in longitudinal text requires distinguishing current affect estimation from future affective change forecasting. Existing approaches often treat each text as an independent observation and apply similar assumptions to both tasks, without testing whether they rely on different information sources. This paper investigates that distinction using longitudinal self-reported ecological essays and feeling-word entries. We propose the Trait--State Affective Prediction (TSAP) framework and its temporal extension E-TSAP for per-text valence and arousal prediction, evaluated on a held-out prediction test set of 1,737 entries from 91 users. We further propose the Affective Change Forecaster Hybrid (ACF-Hybrid) for next-step affective change forecasting, evaluated on a held-out forecasting test set of 46 users. For prediction, E-TSAP achieves composite Pearson correlations of $0.670$ for valence and $0.449$ for arousal. For forecasting, textual representations perform worse than compact numeric trajectory baselines: the text-inclusive model achieves only $r=0.316$ for valence and $r=0.284$ for arousal, whereas a simple prior-state baseline reaches $r=0.615$ and $r=0.670$, respectively. ACF-Hybrid, using dimension-specific numeric trajectory features, achieves $r=0.659$ for valence and $r=0.658$ for arousal. These results show that textual semantics support current affect prediction, whereas future affective change is better captured through prior numeric trajectory dynamics.
\end{abstract}

\begin{keywords}
longitudinal affect modeling \sep
valence--arousal prediction \sep
affect forecasting \sep
personalized emotion modeling \sep
user modeling \sep
temporal emotion dynamics
\end{keywords}

\maketitle

\section{Introduction}
\label{sec:introduction}
Emotion is dynamic and context-dependent, continuously shaped by personal experiences and a person's circumstances. Repeated self-reported text collected over time, such as diaries, self-reflections, and/or journal entries, is not an independent instance of emotions. Rather, each entry reflects a combination of current linguistic expression, relatively stable affective tendencies, and recent emotion history. Longitudinal affect modeling aims to understand and predict affective states from such temporally ordered observations by accounting for both individual differences and emotional change over time. More precisely, affective modeling finds patterns in human emotions across multiple modalities, including text, speech, facial expressions, and physiological signals \citep{picard1997affective}. Despite substantial advances, most text-based affective models still take each document/text as an independent instance. Therefore, such models tend to overlook user-specific characteristics and temporal dependencies that shape emotional experiences. 

Recent user-centered studies show that past emotional states and behavioral trajectories contain useful patterns that cannot be found if observations are considered as independent instances \citep{LIU2023103256, WAN2025103967}.
While past emotional states and behavioral trajectories provide useful context, they do not necessarily contribute equally to all longitudinal affective tasks. Estimating a person's current emotional state from text is fundamentally different from forecasting how that state will evolve in the future. For example, in one user trajectory, the user first writes, I'm good right now. I feel content. Calm. Just trying to get some school work done,'' with valence $v=2$ and arousal $a=0$. Approximately seven hours later, the same user writes, I do not feel well. I keep feeling dizzy,'' with valence $v=-2$ and arousal $a=0$. While the current text may provide sufficient evidence for estimating present affect, forecasting future change requires information about prior emotional trajectories and temporal patterns. Affect prediction aims to estimate current affective states from available text. Affect forecasting, in contrast, seeks to anticipate future affective change from historical observations. Consequently, the information that supports accurate current-state estimation may not necessarily support forecasting future affective change. Current-state prediction primarily depends on linguistic evidence expressed in the text, whereas forecasting relies more heavily on temporal dynamics and prior affective trajectories.

To investigate this problem, we adopt the dimensional valence--arousal (VA) framework \citep{russell1980circumplex, bradley1994measuring}, where valence represents the positive--negative polarity of an emotional state and arousal captures its activation intensity. Unlike discrete emotion categories, continuous affective dimensions preserve subtle variations in emotional experience and allow changes in both magnitude and direction to be modeled over time. These characteristics make the VA framework particularly suitable for longitudinal affect analysis.

Early researches for VA prediction rely on affective lexicons and aggregation methods \citep{warriner2013norms, mohammad2018obtaining} and then recurrent neural architectures incorporating contextual information \citep{matero_autoregressive_2020}. More recently, Large Language Models (LLMs) achieve strong performance on sentence- and document-level affect prediction tasks \citep{mendes2023quantifyingvalencearousaltext, becker_dimstance_2026, lin_enhanced_2024}. But these kind of approaches produces a unique output per text example without taking into consideration any aspect of authorship, time, or emotion history before. Therefore, it is difficult to tell the extent of the contribution made by the language semantics alone as compared to user and temporal context.


This limitation is especially important when affective states are self-reported by the subjects over a long period of time. Most of the challenges in longitudinal affect datasets are not present in conventional benchmark datasets. Self-reported annotations are based on the emotional state experienced by an individual, rather than on the interpretation of an external observer \citep{schneider2025datasetsvalencearousalinference}, and thus create a closer link between language and personal affect. However, user participation varies greatly, observation schedules are frequently inconsistent, and repeated observations reveal both consistent personal traits and dynamic emotional shifts. As a result, models trained on combined observations can achieve good performance by learning user-specific baselines but without accurately modeling the emotional dynamics over time \citep{ganesan_word_2026}. Thus, longitudinal affect modeling needs models and evaluation protocols that decouple stable individual differences from within-individual affective variation over time.

Recent studies provide evidence for both sides of this problem. On one hand, contextual text representations support increasingly accurate affective inference \citep{HUANG2022102822, WANG2023103151}. On the other hand, temporal trajectories support forecasting future emotional states and behavioral outcomes \citep{XI2023103254, WAN2025103967}. Despite these advances, the two directions are typically evaluated independently, making it difficult to determine whether they rely on the same dominant sources of information. Consequently, it remains unclear whether affect prediction and affect forecasting rely on the same dominant information sources or whether different sources of information drive performance in each task.

This paper frames a simple but previously understudied question: \textbf{\textit{Do affect prediction and affect forecasting rely on the same dominant sources of information? If not, which information sources contribute most strongly to each task?}} Answering this question is important because model architectures, feature design, and evaluation protocols are often transferred between prediction and forecasting settings without verifying whether the underlying information requirements are comparable. To the best of our knowledge, a few studies systematically compare affect prediction and affect forecasting within a unified experimental framework. As a result, the relative contributions of textual semantics, user-specific characteristics, and temporal dynamics remain poorly understood. So, it remains unclear whether textual semantics, user-specific characteristics, and temporal dynamics contribute similarly to both tasks or whether their importance changes when the objective shifts from prediction to forecasting. A clearer understanding of these information sources can improve both model design and evaluation for longitudinal affect modeling.

To investigate this question, we use a longitudinal dataset of self-reported ecological essays and feeling-word entries \citep{soni-etal-2026-semeval}. For current affect prediction, we develop the Trait--State Affective Prediction (TSAP) framework and its temporal extension E-TSAP, which estimate valence and arousal from text, user information, and prior affective features. For affect forecasting, we develop the Affective Change Forecaster Hybrid (ACF-Hybrid), which predicts future affective change from compact trajectory-based representations derived from historical observations. All models are evaluated using a protocol that separates within-user and between-user performance, allowing us to distinguish models that capture stable individual baselines from those that genuinely track temporal affective variation.

Across prediction and forecasting experiments, we observe a clear information-source asymmetry between prediction and forecasting. For current affect prediction, textual semantics provide the dominant signal, while user-specific and temporal features contribute smaller but consistent improvements. In contrast, textual representations provide little benefit for forecasting and can even reduce performance relative to compact trajectory-based features. These findings suggest that the information required to estimate current affect differs from the information required to forecast future affective change.

\subsection*{Contributions}
We summarize the contributions of our work as follows:
\begin{itemize}[itemsep=1pt, topsep=2pt]
\item We show that affect prediction and affect forecasting rely on different sources of information. While textual semantics provide the dominant signal for current affect prediction, temporal trajectories play a substantially larger role in forecasting future affective change.

\item We propose the Trait--State Affective Prediction (TSAP) framework and its temporal extension E-TSAP for personalized valence--arousal prediction by integrating textual, user-specific, and temporal affective information.

\item We develop the Affective Change Forecaster Hybrid (ACF-Hybrid) framework for forecasting future affective change and systematically investigate the contribution of textual, user-specific, and trajectory-based information sources.

\item We conduct a comprehensive empirical evaluation across prediction and forecasting settings, including competitive baselines, feature ablations, unseen-user experiments, and decomposed within-user and between-user analyses.

\item We provide a unified experimental framework for analyzing how textual semantics, user-specific characteristics, and temporal dynamics contribute to longitudinal affect modeling under both prediction and forecasting objectives.

\end{itemize}

\subsection*{Organization}
The remainder of the paper is organized as follows. Section~\ref{sec:related_work} reviews prior work on affect prediction, longitudinal affect modeling, and affect forecasting. Section~\ref{sec:dataset_analysis} presents the dataset and its characteristics. Section~\ref{sec:prob_formulation} formalizes the prediction and forecasting tasks. Section~\ref{sec:framework} introduces the proposed frameworks. Section~\ref{sec:experiments} describes the experimental setup. Section~\ref{sec:results} presents the results and analysis. Section~\ref{sec:discussion} discusses the implications of the findings. Section~\ref{sec:limitations} presents the limitations and future research directions. Finally, Section~\ref{sec:conclusion} concludes the paper.

\section{Related Work}\label{sec:related_work}


This section reviews prior work on dimensional emotion representation (Section~\ref{subsec:2.1}), emotion modeling datasets (Section~\ref{subsec:2.2}), text-based valence--arousal prediction (Section~\ref{subsec:2.3}), longitudinal and personalized affect modeling (Section~\ref{subsec:2.4}), and evaluation in user-centered  longitudinal settings (Section~\ref{subsec:2.5}). Contextual language models have steadily improved affect prediction from text, and longitudinal studies consistently find that temporal trajectories and user-specific context carry meaningful predictive signal. But these two directions develop mostly in separate tracks, with little direct comparison between them. Whether current affect prediction and future affective change forecasting draw on the same dominant information sources remains an open question, and that is what this paper directly addresses.


\subsection{Emotion Representation} \label{subsec:2.1}
Dimensional models place emotion in a continuous affective space rather than sorting it into discrete categories \citep{russell1980circumplex}. The valence--arousal (VA) framework remains the most common choice for text-based affect prediction, where valence tracks positive-negative polarity and arousal tracks activation level \citep{bradley1994measuring}. Extensions such as the VAD model add dominance as a third dimension \citep{mehrabian1974basic}, and lexical resources such as the Warriner norms \citep{warriner2013norms} and NRC-VAD \citep{mohammad2018obtaining} provide large-scale affective ratings. However, dominance annotations are less consistently available across text datasets \citep{mendes2023quantifyingvalencearousaltext}. This work therefore adopts the VA framework because it captures gradual variations in emotional experience and naturally supports the analysis of affective trajectories over time.

\subsection{Emotion Modeling Datasets} \label{subsec:2.2}
Emotion corpora cover multimodal laboratory recordings \citep{busso2008iemocap, koelstra2011deap, ringeval2013introducing}, social media and sentence-level text benchmarks \citep{buechel_emobank_2017, preotiuc-pietro-etal-2016-modelling}, and lexical affect resources \citep{warriner2013norms, mohammad2018obtaining}. These datasets have enabled controlled benchmarking, but many treat text instances as independent observations and do not preserve temporal continuity across entries from the same individual. Many also rely on external observer annotations rather than self-reported affect, which can separate the modeled label from the writer's experienced emotional state \citep{schneider2025datasetsvalencearousalinference}. Recent longitudinal datasets provide a complementary perspective by pairing user-indexed text with repeated self-reported affective states \citep{soni-etal-2026-semeval}. Such datasets make it possible to investigate how textual, temporal, and user-specific information contribute to affect modeling over time.

\subsection{Text-Based Valence--Arousal Prediction} \label{subsec:2.3}
Early approaches to affect prediction used psycholinguistic lexicons and classical regression over sparse lexical features \citep{warriner2013norms, mohammad2018obtaining, buechel_emobank_2017}. Neural and transformer-based models substantially improved performance by learning contextual representations from text. \citet{mitsios_improved_2024} showed that incorporating the geometry of the valence--arousal space through ordinal classification can reduce error severity. \citet{mendes2023quantifyingvalencearousaltext} fine-tuned multilingual transformers across 34 datasets in 13 languages, achieving Pearson correlations of $0.810$ for valence and $0.695$ for arousal, while also confirming that arousal is generally harder to predict than valence. Recent work further explores parameter-efficient fine-tuning \citep{lin_enhanced_2024}, LLMs and related dimensional affective tasks \citep{becker_dimstance_2026}, and domain-specific applications such as crisis counseling de-escalation \citep{tripodi-etal-2025-assessing}. Overall, these studies demonstrate that textual representations contain strong signal for estimating current affective states. However, most approaches still generate one prediction per text instance and rarely account for authorship, temporal position, or prior affective history. Consequently, it remains unclear how much additional information can be gained from user-specific and temporal context in longitudinal settings.


\subsection{Longitudinal and Personalized Affect Modeling} \label{subsec:2.4}
Longitudinal affect modeling introduces temporal and user-specific structure that is not captured by independent document-level prediction. \citet{matero_autoregressive_2020} showed that affective dependencies can be modeled from longitudinal language sequences using autoregressive neural architectures. \citet{hills_exciting_2024} introduced time-aware hierarchical transformers with decay mechanisms for irregular posting intervals, while \citet{tseriotou-etal-2023-sequential} used path-signature representations to summarize user histories for personalized longitudinal language modeling. Related work on personalization shows that user identifiers and person-specific representations can improve affective or sentiment prediction \citep{mireshghallah_useridentifier_2022, wortwein_neural_2023}. \citet{soni_evaluation_2025} further showed that human-aware language representations can improve user-centered affective tasks.

These studies confirm that temporal structure and user-specific context carry predictive signal beyond the current text. However, they primarily focus on improving current affect prediction or modeling user trajectories independently. Consequently, the relative contribution of textual semantics, user-specific characteristics, and temporal dynamics remains unclear when the objective shifts from predicting current affect to forecasting future affective change.

\subsection{Evaluation in Longitudinal Settings} \label{subsec:2.5}
Pearson correlation and mean absolute error (MAE) are standard metrics for valence--arousal prediction \citep{mendes2023quantifyingvalencearousaltext, buechel_emobank_2017}. In multimodal affect modeling, the Concordance Correlation Coefficient is also common because it penalizes differences in mean and variance between predicted and observed sequences \citep{wortwein_neural_2023}. In longitudinal user-centered settings, however, pooled metrics can conflate stable between-user differences with within-user affective variation. A model may score well by learning individual baselines while failing to track affective change within the same person. \citet{ganesan_word_2026} demonstrated this problem in longitudinal NLP, and \citet{soni_evaluation_2025} reported substantially different between-user and within-user performance in human-centered affective tasks. These findings suggest that strong overall performance does not necessarily imply effective longitudinal affect modeling. Separating within-user and between-user performance therefore gives a clearer picture of whether a model tracks individual affective dynamics or simply exploits stable differences across users.

\subsection{Research Gap} \label{subsec:2.6}
Prior work demonstrates that textual representations provide strong signals for dimensional affect prediction, while longitudinal studies show that temporal trajectories and user-specific context contribute useful information for modeling emotional dynamics. However, these research directions have largely evolved independently. Affect prediction studies typically evaluate models that estimate current affective state from text, whereas longitudinal forecasting studies focus on future emotional outcomes using historical observations.

As a result, it remains a research gap whether affect prediction and affect forecasting rely on the same dominant sources of information. Existing studies rarely compare textual semantics, user-specific characteristics, and temporal dynamics within a unified experimental framework. Therefore, the relative importance of these information sources across prediction and forecasting tasks remains a question. To address this gap, we conduct a unified empirical investigation of affect prediction and affect forecasting using the same longitudinal dataset and evaluation framework. 

\section{Emotion Modeling Dataset}
\label{sec:dataset_analysis}
This section covers the longitudinal affect dataset and identifies the empirical properties that motivate the proposed modeling design. We analyze dataset structure, affect distributions, user-level variation, temporal dynamics, and linguistic markers. The analysis shows that the data are hierarchical, temporally irregular, and dimension-specific, requiring user-aware and temporally informed modeling rather than independent document-level regression.

\subsection{Dataset Overview}
\label{subsec:dataset_overview}
We use the SemEval-2026 Task~2 \citep{soni-etal-2026-semeval} dataset, which focuses on modeling emotion variation in longitudinal self-reported text. Each entry is paired with the author's own valence and arousal ratings, making the dataset suitable for studying subjective affective states as experienced by the author rather than inferred by external annotators. Each observation includes an anonymized \textit{user\_id}, a unique \textit{text\_id}, raw text, timestamp, collection phase, a binary \textit{is\_words} flag, and self-reported \textit{valence} and \textit{arousal} ranging from $-2$ to $2$, and $0$ to $2$, repspectively. Overall, the data contains 4,501 records: 2,764 in the training set (137 users) and 1,737 in the held-out prediction test split (91 users). 

In Table~\ref{tab:1}, we show example longitudinal sequences for three users (user 17, 50, and 57), demonstrating how affective states change over time ordered entries. The dataset contains two complementary text types: feeling-word entries and narrative essays. Feeling-word entries provide direct affective cues, whereas essays express affect through experiences, events, and reflections. Together, they require models to capture both explicit and contextual affective signals.

\begin{table}
\centering
\caption{Representative longitudinal entries from the affect dataset. Three temporally ordered entries are shown for each user, illustrating how affective states are expressed through both narrative essays and feeling-word entries.}
\label{tab:1}
\scriptsize
\resizebox{\textwidth}{!}{%
\begin{tabular}{p{2.5cm} p{2.2cm} p{8.2cm} c c}
\toprule
\textbf{Date} & \textbf{Text type} & \textbf{Text} & \textbf{Valence} & \textbf{Arousal} \\
\midrule

\multicolumn{5}{c}{\textbf{User 17}} \\
\midrule
2021-06-16 12:36:09 
& Essay 
& I feel okay right now. Of course my shift is fixing to start, so we will see in a few hours how I'm feeling. Hopefully the shift and the people won't be that bad. I'm pretty happy/content at the moment and feeling okay today. 
& 2 & 0 \\

2022-05-01 10:01:00 
& Feeling word 
& Calm, Content, Sluggish, Tired, Serene 
& 2 & 0 \\

2022-05-01 15:15:00 
& Feeling word 
& Tired, Exhausted, Lonely, Sad, Depressed 
& -1 & 0 \\

\midrule
\multicolumn{5}{c}{\textbf{User 50}} \\
\midrule
2021-06-12 17:19:52 
& Essay 
& I've been feeling pretty peaceful and at ease. I'm in the middle of three days off from work and even though I have lots of other responsibilities at the moment they don't feel terribly overwhelming and I feel confident in taking them on. 
& 1 & 0 \\

2022-02-10 15:30:07 
& Feeling word 
& Wired, Excited, Anticipatory 
& 1 & 2 \\

2022-02-11 20:11:32 
& Feeling word 
& Anxious, Waiting, Nervous 
& -1 & 2 \\

\midrule
\multicolumn{5}{c}{\textbf{User 57}} \\
\midrule
2024-10-26 12:00:25 
& Feeling word 
& Calm, Excited, Happy, Calm, Calm 
& 1 & 0 \\

2024-10-26 17:00:22 
& Essay 
& When I came home I felt invigorated after a good gym workout. I cooked and ate lunch. Post meal and shower, I am feeling calm, sleepy and relaxed. I am going to a Halloween party tonight so I am also excited. However, I am anticipating being tired after the party is over. 
& 0 & 0 \\

2024-10-26 22:00:27 
& Feeling word 
& Anticipating, Tired, Excited, Happy, Calm 
& 0 & 1 \\

\bottomrule
\end{tabular}%
}
\end{table}

\subsection{Longitudinal and User-Level Structure}
\label{subsec:dataset_structure}
The dataset has a hierarchical longitudinal structure in which repeated observations are nested within users and ordered over time, as in Table~\ref{tab:1}. Therefore, entries cannot be treated as fully independent instances; each text may reflect the current linguistic content, the user's baseline affective tendency, and the user's recent emotional history.

Participation varies considerably across users. In the training split, users contribute 20.18 entries on average, with a median of 14 and a range from 2 to 206 entries. Twenty-five users, or 18\% of training users, contribute five or fewer entries. The Gini coefficient of 0.51 indicates that a small number of highly active users account for a disproportionate share of the corpus. This imbalance makes user representation and cross-user generalization important concerns.

Temporal spacing is also irregular. Although each user's entries are chronologically ordered, the median inter-observation gap is 6.49 hours, while the standard deviation is 418.87 hours. The data therefore do not form a uniformly sampled time series. These observations indicate that temporal information is potentially useful but cannot be modeled as a regularly sampled process. Effective longitudinal models must therefore leverage available affective history while remaining robust to highly uneven participation patterns and irregular observation intervals.


\subsection{Affect Distribution}
\label{subsec:affect_distribution}
\begin{figure}
    \centering
    \includegraphics[width=\textwidth]{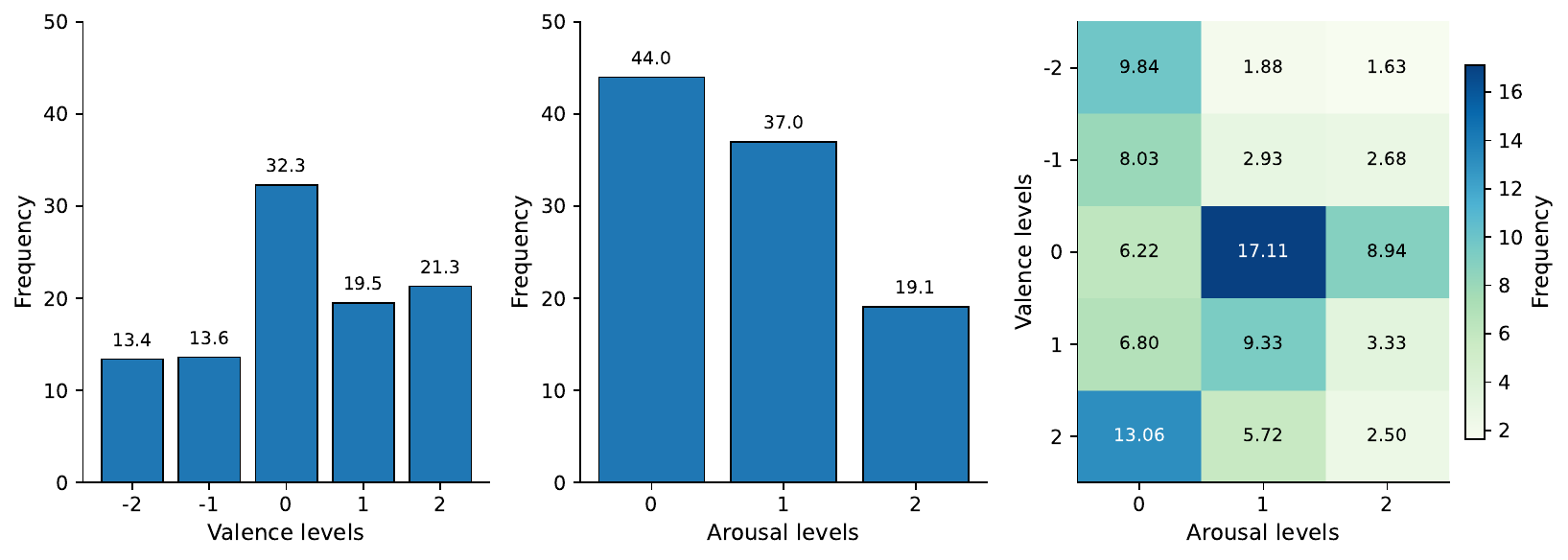}
    \caption{Distribution of valence and arousal labels. The left panel shows the marginal distribution of valence levels, the middle panel shows the marginal distribution of arousal levels, and the right panel shows the joint valence--arousal distribution as the percentage of samples in each label combination.}

    \label{fig:1}
\end{figure}

Valence and arousal are almost linearly independent, with a Pearson correlation of $r=0.026$. The joint distribution is also skewed, as shown in Figure~\ref{fig:1}. Valence and arousal exhibit different distributional properties. Valence is approximately balanced across its five levels, whereas arousal shows a strong concentration at the lowest activation level (44\% of observations). The two dimensions are nearly independent (r = 0.026). These differences suggest that valence and arousal may require different modeling assumptions and evaluation considerations.



\subsection{Between-User and Within-User Variation}
\label{subsec:multilevel_variation}

Affective variation in longitudinal data arises from two complementary sources: (1) stable differences between users and (2) fluctuations within the same user over time. It is important to distinguish these sources because they place different demands on affective models. Models that primarily learn stable user tendencies may achieve strong overall performance without accurately tracking affective change. Conversely, models that capture within-user variation must be sensitive to temporal dynamics beyond user identity. To separate these sources, we compute the Intraclass Correlation Coefficient (ICC) for each affective dimension \citep{shrout1979intraclass}.

Valence has an ICC of 0.327 such that 32.7\% of total variance is due to differences between users that are stable over time and 67.3\% of variance is within users over time. Arousal has a lower ICC of 0.164, meaning that only 16.4\% of its variance is between users and 83.6\% is within users. This difference points to a key distinction between the two affective dimensions. Valence has a large stable user component, indicating there may be useful signal in personalization. In contrast, arousal varies mainly within users over time, making the task less dependent on the user and more on the short-term affective dynamics.

\subsection{Temporal Structure}
\label{subsec:temporal_dynamics}

The temporal structure in the data is examined through affective inertia, volatility, and lag-based dependency. Affective inertia is defined as the tendency of emotional states to persist across consecutive observations \citep{kuppens2010emotional}. Our analysis shows that at the aggregate level, consecutive affective states show moderate short-term association, with lag-1 correlations of $r=0.337$ for valence and $r=0.239$ for arousal. At the individual user level, however, these correlations drop close to zero, reaching $0.004$ for valence and $-0.049$ for arousal.

Emotion trajectories are also highly volatile. On average, valence shifts by $1.026$ and arousal by $0.627$ between consecutive observations, meaning affective states jump considerably from one entry to the next rather than changing gradually. Within-user variability follows the same pattern, with standard deviations of $1.048$ for valence and $0.671$ for arousal. Multi-lag correlations suggest that prior affective states carry some predictive signal, though that signal is limited. For valence, correlations decline from $r=0.337$ at lag 1 to $r=0.292$ at lag 2 and $r=0.281$ at lag 3. Arousal shows a weaker and less consistent pattern, with correlations of $r=0.239$, $r=0.225$, and $r=0.260$ across lags 1 to 3.

These results show that prior affective states provide temporal signal, but that signal is irregular, dimension-specific, and user-dependent. Temporal modeling should therefore not assume smooth or uniformly persistent trajectories. Instead, the analysis motivates compact causal temporal features that capture recent affective history while remaining robust to sparse and heterogeneous user sequences.


\begin{figure}
\centering
    \includegraphics[width=0.88\textwidth]{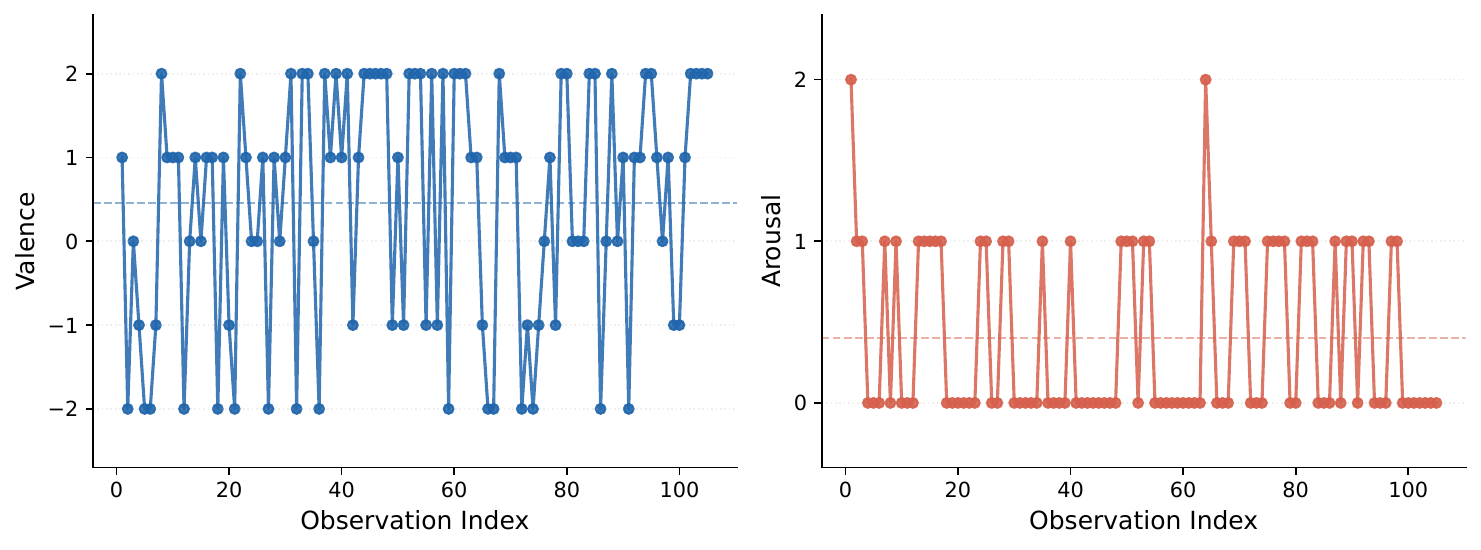}
    \caption{A high-variability user trajectory. Valence and arousal fluctuate rapidly across the observation period, showing why user-aware and temporally informed modeling is needed.}
    \label{fig:2}
\end{figure}

Figure~\ref{fig:2} illustrates rapid movement across the valence--arousal space for a representative user. This variability motivates the distinction between \textit{affect prediction}, which estimates the current affective state from the current text, and \textit{affect forecasting}, which estimates future affective change from prior affective history.

\subsection{Linguistic Markers}
\label{subsec:linguistic_markers}
To identify interpretable textual signals associated with dimensional affect, we compute lexical associations between words and observed valence--arousal scores. Texts are represented using TF-IDF weighted unigram features \citep{salton1988term}, and Pearson correlations are computed between each word feature and each affective dimension across all 4,501 labeled entries for post-evaluation descriptive analysis only. These lexical statistics characterize linguistic patterns in the corpus and are not used for model training, validation, model selection, or test-time inference.

\begin{figure}
    \centering
    \begin{subfigure}{0.48\textwidth}
        \centering
        \includegraphics[width=\linewidth]{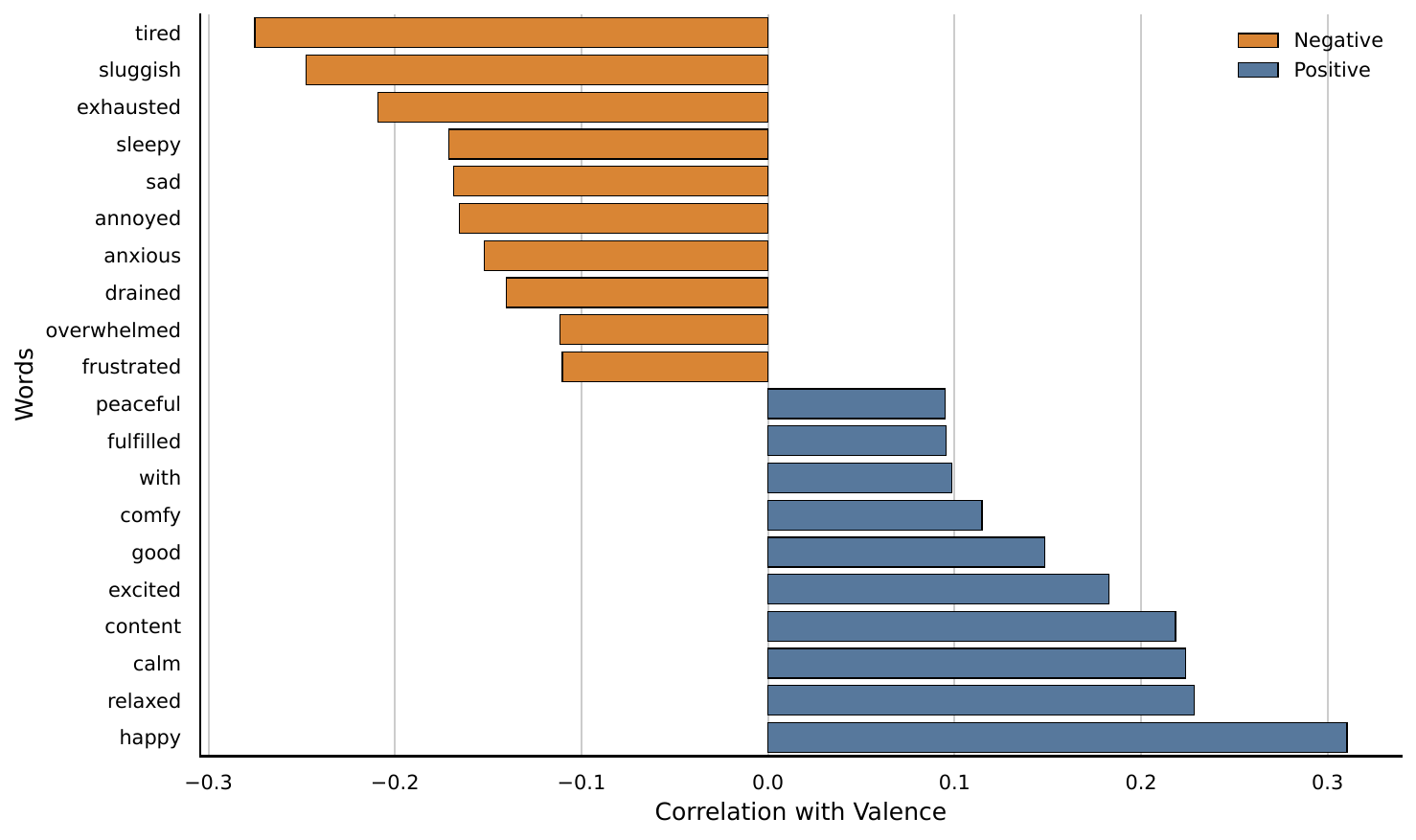}
    \end{subfigure}
    \hfill
    \begin{subfigure}{0.48\textwidth}
        \centering
        \includegraphics[width=\linewidth]{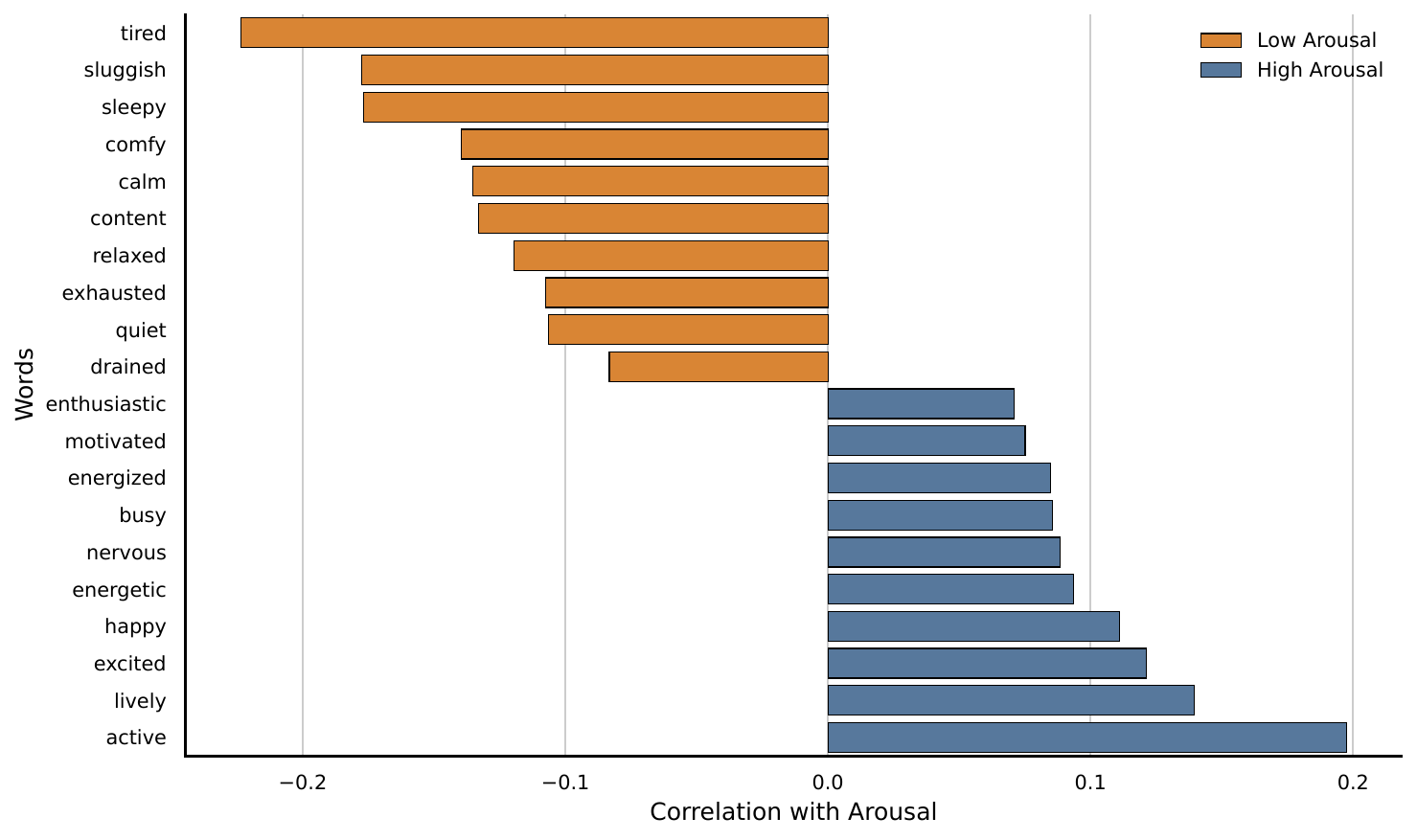}
    \end{subfigure}
    \caption{Top unigrams and their Pearson correlations with valence (left) and arousal (right).}
    \label{fig:3}
\end{figure}

Valence is primarily associated with sentiment-polarity terms. Words such as \textit{happy}, \textit{relaxed}, \textit{calm}, \textit{content}, and \textit{excited} are positively associated with valence, whereas \textit{tired}, \textit{sluggish}, \textit{exhausted}, \textit{sleepy}, \textit{sad}, and \textit{annoyed} are negatively associated. Arousal follows an activation pattern. Words such as \textit{active}, \textit{lively}, \textit{excited}, \textit{energetic}, and \textit{motivated} are positively associated with arousal, while \textit{tired}, \textit{sluggish}, \textit{sleepy}, \textit{calm}, and \textit{relaxed} are negatively associated. These dimension-specific lexical patterns are visualized in Figure~\ref{fig:3}.

This distinction matters because some words have different implications across dimensions. For instance, \textit{calm} and \textit{relaxed} depict positive valence and low arousal. This lexical cue can thus function as an emotional positive cue and as a cue for low-level activation. The patterns presented here show that the meaning of the affect is not equally spread throughout the dimensions. This indicates that lexical cues play different roles in predicting the valence and arousal of words, which justifies modeling them as separate but related targets.

The same trend can be seen in the category-level analysis. Words of positive emotions have higher valence, negative emotion words have lower valence, high arousal words have higher activation, and low arousal words have lower activation. Physical-state terms show negative average valence on the whole, reflecting the tendency for bodily discomfort and fatigue to co-occur with lower affective states.

\subsection{Modeling Implications}
\label{subsec:dataset_modeling_implications}
There are several important properties in the dataset that need to be considered when doing longitudinal affect modeling. First, the inherent nature of the data makes the observations user-dependent and longitudinal, which means that the observations can no longer be considered entirely separate events. Second, user contribution to data generation is extremely uneven, resulting in a scenario where a few active users make the vast majority of contributions. Consequently, any effective model needs to generalize past those users who have contributed a lot rather than just learning from them. Thirdly, valence and arousal have different statistical properties. The between-user element is more dominant in valence, implying that personalized information would be valuable for modeling an individual’s affective state. On the other hand, within-user variability is highly significant in arousal, which is also marked by a low activation threshold effect. These differences show that the two concepts cannot be expected to share similar behavior in a model evaluation process. Fourth, temporal history contains predictive information yet does not follow the typical patterns of time series data. User activity is inconsistent, observation timespans differ greatly, and affective profiles exhibit high volatility. Hence, methods relying on temporal modeling may overlook the heterogeneity carried by the dataset. Rather, temporal information needs to be taken into account such that it is resilient to sparseness and irregularity in users' activities. Finally, regarding lexical analysis, it turns out that affective information can be extracted from the present text. Valence is connected with sentiment, while arousal is connected with activation or energy. Nevertheless, various properties that are related to other dimensions indicate that affective information goes beyond a single emotional dimension.

Altogether, these analyses uncover a dataset where textual content, user-specific factors, and temporal dynamics all influence affective variation, but in different ways. While current affective states still dominate the text, the temporal history captures aspects of affective continuity that are not directly observable from the current text alone. These observations motivate our central hypothesis that affect prediction and affect forecasting might depend on different dominant information sources despite operating on the same longitudinal user trajectories.

\section{Task Formulation}
\label{sec:prob_formulation}
In this work, we study two related but different longitudinal affect modeling tasks: affect prediction and affect forecasting. Both tasks operate on the same user trajectories and affective labels, but have different objectives, input signals, and information sources. Affect prediction estimates the user’s current affective state based on the current text, while affect forecasting estimates the future affective change from prior affective history. This section formalizes both tasks.
\subsection{Notation}
\label{subsec:notation}

Let $\mathcal{U} = \{u_1, u_2, \ldots, u_M\}$ denote the set of $M$ users. For each user $u \in \mathcal{U}$, we observe a temporally ordered sequence of $T_u$ observations:
\begin{equation}
\mathcal{S}_u =
\bigl\{(x_{u,i}, \mathbf{y}_{u,i}, t_{u,i})\bigr\}_{i=1}^{T_u}.
\label{eq:sequence}
\end{equation}

Here, $x_{u,i}$ denotes the textual entry, $t_{u,i}$ denotes the associated timestamp, and $\mathbf{y}_{u,i} = (v_{u,i}, a_{u,i}) \in \mathbb{R}^2$ denotes the affective state consisting of valence $v_{u,i} \in [-2,2]$ and arousal $a_{u,i} \in [0,2]$. All observations are chronologically ordered such that $t_{u,i} < t_{u,i+1}$ for consecutive observations.


\subsection{Affect Prediction}
\label{subsec:prediction}

The affect prediction task estimates the current affective state of a user from the semantic content of the current textual entry. Given user $u$ and text $x_{u,i}$ at observation $i$, the objective is to learn the mapping:
\begin{equation}
f_{\text{pred}}(x_{u,i}, u)
\rightarrow
\hat{\mathbf{y}}_{u,i},
\label{eq:prediction_mapping}
\end{equation}
where $\hat{\mathbf{y}}_{u,i}=(\hat{v}_{u,i},\hat{a}_{u,i})$ denotes the predicted valence and arousal scores.

The prediction target is the absolute affective state $\mathbf{y}_{u,i}=(v_{u,i},a_{u,i})$ at the same observation. The model has access to the current text and user identity, but not to future observations. This task is formulated as direct regression over current affective states rather than affective change. The objective is to estimate the user's affective state from the information available at the same observation. The semantics of the text components provide the primary signal, while user-level information reflects stable affective tendencies that are observed across the text components.

\subsection{Affect Forecasting}
\label{subsec:forecasting}
Affect forecasting takes a different starting point. Rather than reading the current text, the task asks how a user's emotional state is likely to shift at the next observation, drawing on what has already been recorded. Prior affective history is therefore the main input, not what the user just wrote. Whether adding text on top of that history actually helps is tested separately through text-inclusive forecasting variants.

For each user trajectory, the affective change between consecutive observations is defined as:
\begin{equation}
\Delta_{u,i}
=
\left(\Delta^{(v)}_{u,i}, \Delta^{(a)}_{u,i}\right),
\label{eq:delta_vector}
\end{equation}
where
\begin{equation}
\Delta^{(v)}_{u,i}
=
v_{u,i+1}-v_{u,i},
\qquad
\Delta^{(a)}_{u,i}
=
a_{u,i+1}-a_{u,i}.
\label{eq:delta_components}
\end{equation}
The final observation in each user sequence has no subsequent observation and therefore does not produce a forecasting target.

Given the affective history up to observation $i$:
\begin{equation}
\mathcal{H}_{u,i}
=
\left\{(v_{u,j},a_{u,j})\right\}_{j=1}^{i},
\label{eq:affective_history}
\end{equation}
the forecasting objective is to learn the mapping:
\begin{equation}
f_{\text{fore}}(\mathcal{H}_{u,i})
\rightarrow
\hat{\Delta}_{u,i},
\label{eq:forecasting_mapping}
\end{equation}
where $\hat{\Delta}_{u,i}=(\hat{\Delta}^{(v)}_{u,i},\hat{\Delta}^{(a)}_{u,i})$ denotes the predicted next-step change in valence and arousal.

The task is set up as continuous regression over future affective change, not over direct estimation of the current affective state. This puts the weight on temporal structure and recent emotional history, which makes forecasting a fundamentally different problem from reading affect out of the current text.

\subsection{Comparison of Prediction and Forecasting}
\label{subsec:task_comparison}

Affect prediction and affect forecasting share the same longitudinal data structure, but they differ in input signals, targets, and modeling assumptions. Table~\ref{tab:2} summarizes these differences, motivating the separate frameworks introduced in Section~\ref{sec:framework}.

\begin{table}[t]
\centering
\caption{Comparison of affect prediction and affect forecasting. Although both tasks operate on the same longitudinal user trajectories, they differ in their objectives, prediction targets, and dominant sources of information.}
\label{tab:2}

\small
\renewcommand{\arraystretch}{1.15}

\begin{tabular}{p{3.0cm} p{5.2cm} p{5.2cm}}
\toprule
\textbf{Aspect} &
\textbf{Affect Prediction} &
\textbf{Affect Forecasting} \\
\midrule

Objective
&
Estimate the user's current affective state
&
Estimate future affective change
\\

Input
&
Current text $x_{u,i}$ and user identity $u$
&
Affective history $\mathcal{H}_{u,i}$
\\

Target
&
Current affective state $\mathbf{y}_{u,i}$
&
Future affective change $\Delta_{u,i}$
\\

Primary information source
&
Textual semantics and user-specific tendencies
&
Temporal dynamics and prior affective states
\\

Prediction level
&
Current observation
&
Next observation
\\

Main challenge
&
Personalization and user heterogeneity
&
Affective volatility and temporal uncertainty
\\

\bottomrule
\end{tabular}
\end{table}


\section{Proposed Framework}
\label{sec:framework}
This section presents the proposed framework for the two tasks defined in Section~\ref{sec:prob_formulation}. For affect prediction, we introduce the Trait--State Affective Prediction (TSAP) framework and its temporal extension E-TSAP. For affect forecasting, we introduce the Affective Change Forecaster Hybrid (ACF-Hybrid), which models future affective change from temporal trajectory features. The design of each component follows the dataset properties identified in Section~\ref{sec:dataset_analysis}.

\subsection{Affect Prediction}
\label{subsec:framework_affect_prediction}

Affect prediction estimates the current valence and arousal of a user from the semantic content of the current textual entry, conditioned on user-level information. The dataset analysis in Section~\ref{sec:dataset_analysis} indicates that this task requires contextual semantic modeling, user-level personalization, and sensitivity to the different statistical properties of valence and arousal. These requirements are addressed through TSAP and its temporal extension, E-TSAP.

\subsubsection{Trait--State Affective Prediction (TSAP)}
\label{subsubsec:tsap}

\begin{figure}
    \centering
    \includegraphics[width=0.92\textwidth]{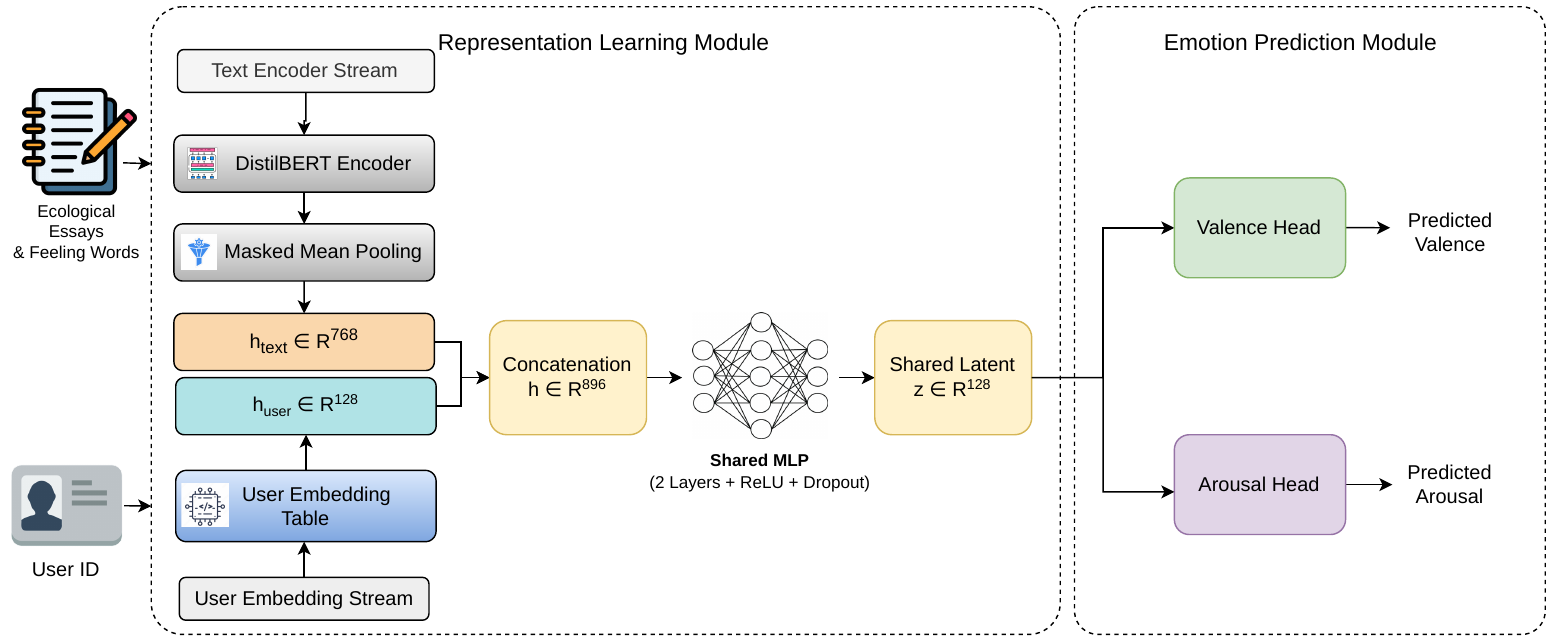}
    \caption{Overview of the Trait--State Affective Prediction (TSAP) framework. Contextual text representations from DistilBERT are combined with learned user embeddings through shared latent representation learning, followed by separate valence and arousal prediction heads.}
    \label{fig:4}
\end{figure}

TSAP is a personalized transformer-based architecture that combines contextual text representations with learned user embeddings. The design draws on the trait--state perspective from personality psychology \citep{steyer1999latent}, which holds that who a person is in a stable sense influences how they react to what is happening around them in the moment. In this setting, user embeddings represent stable affective tendencies, while text representations capture the current affective state expressed in the entry. Figure~\ref{fig:4} shows the overall TSAP architecture.

\paragraph{Text encoder.}
Each textual entry $x_{u,i}$ is encoded with a fine-tuned DistilBERT encoder \citep{sanh2020distilbertdistilledversionbert}. We use DistilBERT because it matched the performance of both BERT and RoBERTa in our encoder comparison experiments while being considerably cheaper to run. The resulting token representations are collapsed into a fixed-length sentence vector through masked mean pooling:
\begin{equation}
\mathbf{h}_{\text{text}}
=
\frac{\sum_{j=1}^{L} m_j \mathbf{h}_j}{\sum_{j=1}^{L} m_j}
\in \mathbb{R}^{768},
\label{eq:mean_pooling}
\end{equation}
where $\mathbf{h}_j$ denotes the hidden representation at token position $j$, $m_j$ denotes the attention mask, and $L$ denotes the sequence length. Mean pooling is selected because it provides stable validation performance across valence and arousal.

\paragraph{User embedding.}
Each user is associated with a learnable embedding vector:
\begin{equation}
\mathbf{h}_{\text{user}} \in \mathbb{R}^{128},
\label{eq:user_embedding}
\end{equation}
representing stable user-level affective tendencies. Users absent during training are mapped to a shared unknown-user embedding, allowing the model to handle unseen-user cases without creating user-specific parameters at inference time.

\paragraph{Fusion and regression.}
The text representation and user embedding are concatenated to form:
\begin{equation}
\mathbf{h}_{\text{TSAP}}
=
\left[
\mathbf{h}_{\text{text}};
\mathbf{h}_{\text{user}}
\right],
\label{eq:tsap_fusion}
\end{equation}
which is passed through a shared feedforward network to obtain a latent representation $\mathbf{z}$. Concatenation is used because fusion comparisons show that gating variants do not provide consistent gains over simple concatenation. The model then uses two independent regression heads:
\begin{equation}
\hat{v}_{u,i}
=
\mathbf{W}_v \mathbf{z} + b_v,
\qquad
\hat{a}_{u,i}
=
\mathbf{W}_a \mathbf{z} + b_a,
\label{eq:tsap_heads}
\end{equation}
to generate valence and arousal predictions. Separate heads are used because valence and arousal show weak empirical association in Section~\ref{subsec:affect_distribution} and differ in their distributional properties. All encoder and prediction-head parameters are fine-tuned jointly.

\paragraph{Training.}
TSAP is trained using a joint mean squared error objective over valence and arousal predictions:
\begin{equation}
\mathcal{L}_{\text{TSAP}}
=
\frac{1}{N}
\sum_{u\in\mathcal{U}}
\sum_{i=1}^{T_u}
\left[
(\hat{v}_{u,i}-v_{u,i})^2
+
(\hat{a}_{u,i}-a_{u,i})^2
\right],
\label{eq:tsap_loss}
\end{equation}
where $N=\sum_{u}T_u$ denotes the total number of training instances. Model selection is based on the composite Pearson correlation metric $r_{\text{comp}}$ computed on the validation set.

\subsubsection{Extended TSAP (E-TSAP)}
\label{subsubsec:etsap}

\begin{figure}
    \centering
    \includegraphics[width=0.92\textwidth]{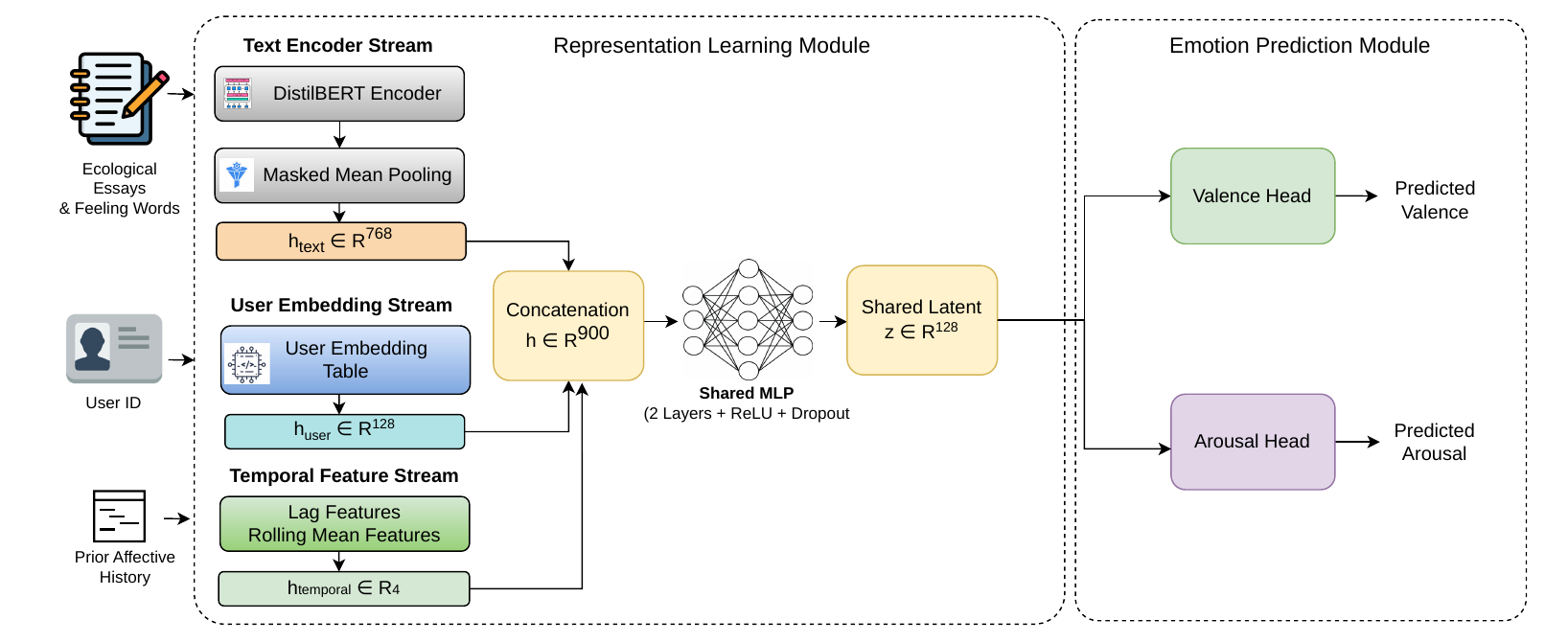}
    \caption{Overview of the Extended Trait--State Affective Prediction (E-TSAP) framework integrating textual, user-level, and temporal affective representations.}
    \label{fig:5}
\end{figure}

E-TSAP extends TSAP by adding a third input stream that represents short-term temporal context from prior affective observations. This extension is motivated by the short-range temporal dependencies identified in Section~\ref{subsec:temporal_dynamics}, which suggest that recent affective states provide complementary information beyond textual semantics and user-level personalization. All TSAP components are retained; only the fusion stage is expanded to include temporal features. The overall E-TSAP architecture is shown in Figure~\ref{fig:5}.

\paragraph{Temporal feature construction.}
Temporal features are computed from prior affective observations using strictly causal aggregation to prevent future information leakage. Lag-based features encode the immediately preceding affective state:
\begin{equation}
\mathbf{f}_{\text{lag},u,i}
=
\left[
v_{u,i-1},
a_{u,i-1}
\right].
\label{eq:lag_features}
\end{equation}
Rolling features summarize recent affective trends:
\begin{equation}
\mathbf{f}_{\text{roll},u,i}
=
\left[
\frac{1}{k}\sum_{j=1}^{k}v_{u,i-j},
\frac{1}{k}\sum_{j=1}^{k}a_{u,i-j}
\right],
\qquad
k=\min(3,i-1).
\label{eq:rolling_features}
\end{equation}
The lag-based and rolling features are concatenated to form:
\begin{equation}
\mathbf{h}_{\text{temporal}}
=
\left[
\mathbf{f}_{\text{lag},u,i};
\mathbf{f}_{\text{roll},u,i}
\right].
\label{eq:temporal_representation}
\end{equation}

All temporal features are computed after chronological sorting within each user sequence, using shifted operations so that the current observation is never included in its own temporal representation. Missing values at sequence boundaries are zero-imputed.

\paragraph{E-TSAP architecture.}
The temporal representation is concatenated with the textual and user-level representations:
\begin{equation}
\mathbf{h}_{\text{E-TSAP}}
=
\left[
\mathbf{h}_{\text{text}};
\mathbf{h}_{\text{user}};
\mathbf{h}_{\text{temporal}}
\right].
\label{eq:etsap_fusion}
\end{equation}
All downstream components remain identical to TSAP, including the shared latent representation and separate valence and arousal regression heads.

\subsection{Affect Forecasting}
\label{subsec:framework_affect_forecasting}

Unlike affect prediction, which estimates the current affective state from text, affect forecasting estimates the next affective change from prior emotion states. The forecasting framework therefore treats numeric affective history as the primary signal and evaluates whether compact trajectory features can better capture next-step affective change than text-inclusive representations.

\subsubsection{Affective Change Forecaster Hybrid (ACF-Hybrid)}
\label{subsubsec:acf_hybrid}

\begin{figure}
    \centering
    \includegraphics[width=\textwidth]{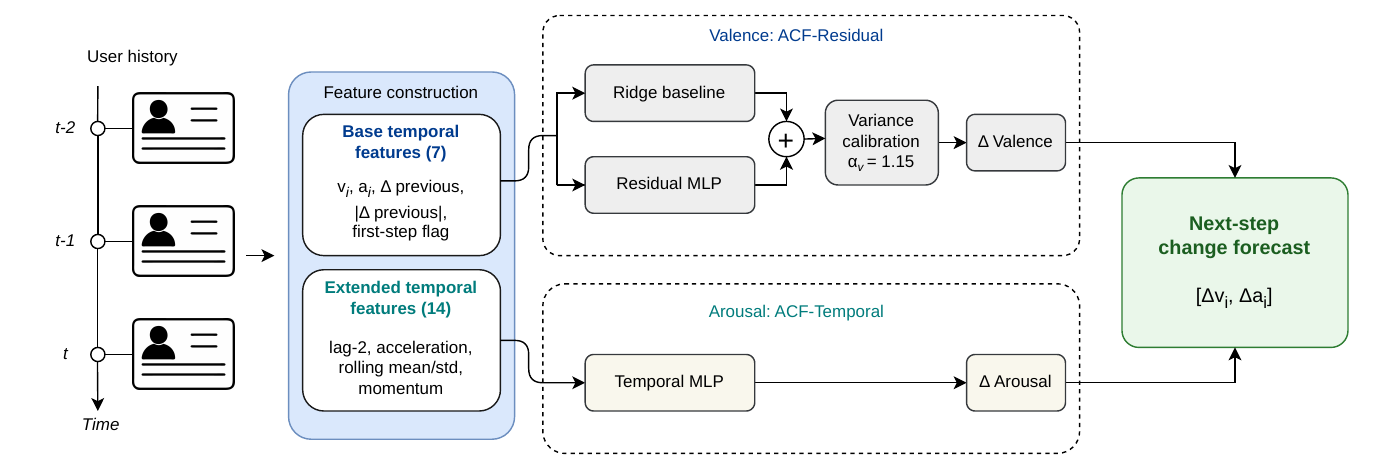}
    \caption{Overview of the ACF-Hybrid framework. ACF-Residual forecasts valence change from base temporal features using a Ridge baseline plus residual MLP, while ACF-Temporal forecasts arousal change from an extended feature set. The two outputs combine into the final next-step forecast $[\Delta v_i, \Delta a_i]$.}
    \label{fig:6}
\end{figure}

ACF-Hybrid is the proposed framework for forecasting next-step affective change. It combines two dimension-specific numeric forecasters: ACF-Residual for valence and ACF-Temporal for arousal. This design reflects the different temporal and distributional properties of the two affective dimensions. Valence benefits from residual correction over a strong linear autoregressive signal, whereas arousal benefits from a richer temporal feature representation. The overall ACF-Hybrid architecture is shown in Figure~\ref{fig:6}.

\paragraph{Base temporal features.}
Forecasting features are computed from prior affective observations using strictly causal construction to prevent future information leakage. The base feature vector $\mathbf{f}_{\text{base},u,i}$ includes the current affective state $(v_{u,i},a_{u,i})$, previous affective changes $(\Delta^{(v)}_{u,i-1},\Delta^{(a)}_{u,i-1})$, absolute previous changes as volatility indicators, and a cold-start indicator for the first observation in a user trajectory.

\paragraph{ACF-Residual for valence.} ACF-Residual breaks valence forecasting into two steps. A Ridge regression model first fits the dominant autoregressive trend: 

\begin{equation} \hat{\Delta}^{(v)}_{\text{linear}} = \boldsymbol{\beta}^{\top}\mathbf{f}_{\text{base},u,i} + \beta_0. \label{eq:acf_linear} \end{equation} 

A small MLP then picks up whatever nonlinear structure the Ridge model leaves behind. The final valence forecast is: 
\begin{equation} \hat{\Delta}^{(v)}_{R} = \hat{\Delta}^{(v)}_{\text{linear}} + \mathrm{MLP}_{\text{res}} \left( \mathbf{f}_{\text{base},u,i} \right). \label{eq:acf_residual} 
\end{equation} 

This decomposition preserves the strong linear trajectory signal while allowing local nonlinear correction.

\paragraph{ACF-Temporal for arousal.}
ACF-Temporal forecasts arousal change using an extended temporal representation. In addition to the base temporal features, it incorporates second-order lag terms, rolling statistics, acceleration features, and direction-consistency indicators. These features capture short-term momentum and local trajectory dynamics that are useful for arousal forecasting. The resulting extended feature vector is mapped to future arousal change through a two-layer MLP.

\paragraph{Hybrid output.}
The final ACF-Hybrid output combines the dimension-specific forecasts:
\begin{equation}
\hat{\Delta}_{\text{Hybrid}}
=
\left(
\hat{\Delta}^{(v)}_{R},
\hat{\Delta}^{(a)}_{T}
\right),
\label{eq:acf_hybrid_output}
\end{equation}
where $\hat{\Delta}^{(v)}_{R}$ is produced by ACF-Residual and $\hat{\Delta}^{(a)}_{T}$ is produced by ACF-Temporal. Unlike TSAP and E-TSAP, the final forecasting framework does not use text representations or user embeddings; it relies on numeric affective history to model next-step affective change. 

Table~\ref{tab:3} summarizes the task, input signal, temporal context, and architecture of each proposed framework component. It also highlights a key distinction between the proposed frameworks. Prediction models are primarily driven by textual representations augmented with user and temporal context, whereas forecasting models rely predominantly on temporal trajectory features derived from prior affective history.

\begin{table}
    \centering
    \caption{Summary of the proposed framework components.}
    \label{tab:3}
    \renewcommand{\arraystretch}{1.20}
    \resizebox{\textwidth}{!}{%
    \begin{tabular}{lllll}
        \hline
        \textbf{Component} & \textbf{Task} & \textbf{Input} & \textbf{Temporal context} & \textbf{Architecture} \\
        \hline
        TSAP
        & Prediction
        & Text + user embedding
        & None
        & DistilBERT + MLP \\

        E-TSAP
        & Prediction
        & Text + user embedding
        & Lag + rolling features
        & DistilBERT + MLP \\

        ACF-Residual
        & Forecasting
        & Numeric affect history
        & Base temporal features
        & Ridge + residual MLP \\

        ACF-Temporal
        & Forecasting
        & Numeric affect history
        & Extended temporal features
        & Temporal MLP \\

        ACF-Hybrid
        & Forecasting
        & Numeric affect history
        & Dimension-specific temporal features
        & ACF-Residual + ACF-Temporal \\
        \hline
    \end{tabular}%
    }
\end{table}

\section{Experiments}
\label{sec:experiments}

This section describes the experimental setup for affect prediction and affect forecasting, including data splits, task-specific input construction, augmentation settings, evaluated models, evaluation metrics, and implementation details.

\subsection{Data Splits}
\label{subsec:data_splits}

We use user-aware and temporally ordered splits to preserve the longitudinal structure of the dataset and evaluate both within-user temporal generalization and cross-user generalization.

\subsubsection{Prediction Splits}
For affect prediction, users with fewer than six observations are excluded before creating the development split to ensure sufficient temporal context for within-user evaluation, resulting in 2,677 instances. For seen users, observations are ordered chronologically, with the earliest 80\% assigned to training and the remaining 20\% assigned to validation. A separate group of unseen users is held out entirely for validation, with no user overlap with the training set. This yields 1,680 training instances and 997 validation instances. The held-out prediction test split of 1,737 instances remains unseen during training and validation and is used only for final evaluation.

\subsubsection{Forecasting Splits}
For affect forecasting, the full 2,764-instance training file is used to construct next-step state-change targets. This yields 2,627 labeled transition rows and 137 unlabeled final rows, one per user sequence. The last labeled transition for each user is held out for validation, giving 137 validation rows that match the one-prediction-per-user structure used at test time. The remaining 2,490 labeled rows form the inner training set. Seven users with only one labeled transition appear only in validation. The held-out forecasting test set contains one target per user for 46 users and is used only for final evaluation.

\subsection{Task-Specific Input Construction}
\label{subsec:task_implementation}

Prediction and forecasting share the same longitudinal dataset but differ in input construction and target definition. For affect prediction, each input consists of the current text $x_{u,i}$ and user identity $u$, and the target is the current affective state $(v_{u,i},a_{u,i})$. For affect forecasting, each input is constructed from prior affective history, and the target is the signed next-step change $(\Delta_{u,i}^{(v)},\Delta_{u,i}^{(a)})$. Forecasting predictions are clipped to the valid transition ranges $[-4,4]$ for valence and $[-2,2]$ for arousal, matching the annotation scales.

Text is used as the primary input for prediction models. In forecasting, numeric temporal features are the primary input. A text-inclusive ACF variant is evaluated only to test whether the text-based TSAP/E-TSAP architecture transfers to future affective change forecasting.

\subsection{Data Augmentation and Transfer Learning}
\label{subsec:data_augmentation}

We evaluate two auxiliary augmentation strategies for affect prediction. First, we use EmoBank as an intermediate valence-arousal supervision source, using writer-perspective annotations and rescaling them to match the SemEval label ranges. This setting tests whether external dimensional affect supervision improves prediction in the target dataset. Second, we use LLaMA-3.1-based paraphrase augmentation following the general motivation of text augmentation with large language models \citep{dai2025auggpt}. The original 2,764 training instances are expanded through paraphrasing and cleaned to remove malformed outputs, yielding 8,282 augmented training instances.

These strategies are treated as auxiliary experiments rather than core components of the proposed framework. They test whether additional lexical variation improves affect prediction, but they do not add new users, timestamps, or true user-specific affective trajectories. Their results are therefore interpreted as evidence about lexical transfer and augmentation limits rather than temporal supervision. Full details are reported in Appendix~\ref{app:data_augmentation}.

\subsection{Models}
\label{subsec:models}

This subsection summarizes the baseline models and proposed architectures evaluated for affect prediction and affect forecasting.

\subsubsection{Prediction Models}
\label{subsubsec:pred_baselines}

For affect prediction, we evaluate baselines spanning statistical, lexical, transformer-based, temporal, and sequential neural approaches. \textbf{GlobalMean} predicts constant global mean valence and arousal values computed from the training set. \textbf{UserMean} predicts user-specific average affective scores, with unseen users assigned the global mean. \textbf{TF-IDF--Ridge} applies $\ell_2$-regularized Ridge regression to TF-IDF weighted unigram and bigram features with a vocabulary size of 10{,}000. \textbf{DistilBERT-Ridge} applies Ridge regression to frozen DistilBERT sentence embeddings obtained through mean pooling. \textbf{DistilBERT-Regressor} fine-tunes DistilBERT with a two-layer MLP regression head, providing a strong text-only neural baseline. \textbf{MovingAvg} ($k=3$) predicts affect as the average of the three most recent prior observations. \textbf{Hybrid (MA+UM)} combines moving-average and user-mean predictions using a weighted combination with $\alpha=0.7$. \textbf{DistilBERT-LSTM} processes chronologically ordered DistilBERT embeddings over fixed-length windows ($k=3$) using an LSTM, evaluating sequential modeling of semantic trajectories.

The proposed prediction models are \textbf{TSAP}, which combines contextual text representations with learned user embeddings, and \textbf{E-TSAP}, which extends TSAP with causal lag-based and rolling temporal features.

\subsubsection{Forecasting Models}
\label{subsubsec:forecast_models}

For affect forecasting, we evaluate conventional baselines and progressively refined variants of the proposed ACF framework. \textbf{RandZero} always predicts zero affective change, representing complete affective inertia. \textbf{Linear-BERT} applies Ridge regression to DistilBERT embeddings of the current text without prior affective information. \textbf{Linear-BERT+Prev} augments textual embeddings with the previous affective state, testing whether textual and numeric signals are complementary. \textbf{Linear-Prev} applies Ridge regression using only the previous affective state and serves as a minimal autoregressive baseline.

The ACF variants isolate the contribution of text, user information, loss function, temporal feature design, and dimension-specific modeling. \textbf{ACF} transfers the E-TSAP-style text-inclusive architecture to affective state-change prediction using textual, user-level, and temporal inputs. \textbf{ACF-Numeric} removes the text encoder and uses numeric temporal features with a compact user embedding. \textbf{ACF-Numeric+} is an auxiliary numeric variant using a hybrid SmoothL1--Pearson objective; implementation details are provided in Appendix~\ref{app:forecasting_variants}. \textbf{ACF-AR} removes user embeddings, yielding a purely autoregressive temporal model. \textbf{ACF-Residual} decomposes valence forecasting into a Ridge-regression baseline and a neural residual corrector. \textbf{ACF-Temporal} uses extended temporal features, including higher-order lags, acceleration, rolling statistics, and direction-consistency indicators. Finally, \textbf{ACF-Hybrid} combines ACF-Residual for valence and ACF-Temporal for arousal as the final forecasting framework.

\subsection{Evaluation}
\label{subsec:evaluation}

We assess the model performance using Pearson correlation and mean absolute error (MAE). For affect prediction, our main metric is the composite Pearson correlation, a Fisher-$z$ average of within- and between-user correlations:
\begin{equation}
r_{\text{comp}}
=
\tanh
\left(
\frac{
\tanh^{-1}(r_{\text{within}})
+
\tanh^{-1}(r_{\text{between}})
}{2}
\right).
\label{eq:r_comp}
\end{equation}
The within-user correlation $r_{\text{within}}$ measures how well predictions track variation within each user over time, while the between-user correlation $r_{\text{between}}$ measures how well predictions distinguish users by their average affective levels. Users with fewer than two valid observations or zero target variance are excluded from within-user correlation computation.

MAE is reported as a complementary error-based metric. For prediction, we report within-user MAE, between-user MAE, and their average:
\begin{equation}
\text{MAE}_{\text{comp}}
=
\frac{1}{2}
\left(
\text{MAE}_{\text{within}}
+
\text{MAE}_{\text{between}}
\right).
\label{eq:mae_comp}
\end{equation}
All metrics are computed separately for valence and arousal.

For forecasting, evaluation is performed at the user level, with one prediction generated per test-user trajectory. Pearson correlation and MAE are reported separately for valence and arousal change. Because the forecasting test set contains 46 users, forecasting correlations are interpreted as task-level performance indicators rather than fine-grained estimates of individual model differences.

\subsection{Training and Implementation Details}
\label{subsec:implementation}

TSAP and E-TSAP are implemented in PyTorch using the Hugging Face Transformers library. The text encoder is initialized from \texttt{distilbert-base-uncased} and fine-tuned during training. Input texts are tokenized with a maximum sequence length of 256 tokens. User embeddings have dimensionality 128, with a dedicated \texttt{UNK} index reserved for unseen users. The shared feedforward network projects the fused representation to 256 dimensions, followed by ReLU activation, dropout ($0.2$), and a second linear layer reducing the representation to 128 dimensions. Optimization uses AdamW with learning rate $2 \times 10^{-5}$, weight decay $0.01$, batch size 16, and early stopping with patience 3 over a maximum of 20 epochs.

Forecasting models are implemented using scikit-learn and PyTorch. The Ridge component in ACF-Residual uses scikit-learn Ridge regression with default regularization strength ($\alpha=1.0$). The residual corrector and temporal MLP are compact feedforward networks using LayerNorm, nonlinear activation, dropout, and one-dimensional output heads. The residual corrector uses smaller hidden layers, while ACF-Temporal uses a wider first hidden layer to support richer temporal features. Temporal features are standardized using a \texttt{StandardScaler} fitted only on the training split, and missing values at sequence boundaries are imputed with zero. ACF-Residual applies mild post-hoc variance calibration ($\alpha_v=1.15$) to valence predictions. A fixed random seed of 42 is used throughout all experiments, which are conducted on an NVIDIA T4 GPU via Google Colab. Table~\ref{tab:4} summarizes the main experimental settings for the prediction and forecasting tasks.

\begin{table}
    \centering
    \caption{Summary of experimental settings for prediction and forecasting tasks.}
    \label{tab:4}
    \renewcommand{\arraystretch}{1.20}
    \begin{tabular}{lll}
        \hline
        \textbf{Setting} & \textbf{Prediction} & \textbf{Forecasting} \\
        \hline
        Input representation & Text + user information & Numeric trajectory features \\
        Text encoder & Fine-tuned DistilBERT & Used only in text-inclusive ACF \\
        User information & User embedding ($\mathbb{R}^{128}$) & Varies by model \\
        Optimizer & AdamW ($2\times10^{-5}$) & Ridge / AdamW \\
        Batch size & 16 & 32--64 for neural numeric models; closed-form fitting for Ridge baselines \\
        Early stopping & Patience 3 & Patience 5 \\
        Primary metric & $r_{\text{comp}}$ & Pearson $r$ \\
        Test set & 1,737 instances & 46 users \\
        Random seed & 42 & 42 \\
        \hline
    \end{tabular}
\end{table}

\section{Results and Analysis}
\label{sec:results}

This section reports the test-set results for the two tasks studied in this paper: affect prediction and affect forecasting. Affect prediction evaluates how well models estimate the current valence and arousal scores of each text entry, while affect forecasting evaluates how well models predict next-step changes in valence and arousal from prior affective history. The results are organized accordingly. Sections~\ref{subsec:results_affect_prediction} and~\ref{subsec:results_affect_forecasting} present prediction and forecasting results, including baseline comparison, ablation, augmentation, and error analysis, respectively. 

\subsection{Affect Prediction}
\label{subsec:results_affect_prediction}

The test set results of all models proposed for affect prediction are presented in Table~\ref{tab:5}. It can be seen that there are three prominent trends in the results obtained. First, global or user-level averages do not suffice as statistical baselines for longitudinal affect prediction. GlobalMean performs poorly on both dimensions, while UserMean improves between-user correlation but results in zero within-user correlation. This suggests that stable user averages are not able to track affective variation in the same individual over time.

\begin{table}
\centering
\caption{Test-set prediction performance across baseline and proposed models. $r_{\text{comp}}$ denotes composite Pearson correlation, $r_{\text{between}}$ between-user correlation, and $r_{\text{within}}$ within-user correlation. MAE$_{\text{comp}}$ denotes composite mean absolute error. Bold indicates the best value per metric and affective dimension. $\dagger$ denotes proposed models. Dashes indicate undefined composite MAE arising from degenerate constant predictions (GlobalMean) or near-zero within-user prediction variance (DistilBERT-LSTM).}
\label{tab:5}
\renewcommand{\arraystretch}{1.15}
\resizebox{\textwidth}{!}{%
\begin{tabular}{lcccccc|cccccc}
\hline
& \multicolumn{6}{c|}{\textbf{Valence (V)}} 
& \multicolumn{6}{c}{\textbf{Arousal (A)}} \\
\cmidrule(lr){2-7} \cmidrule(lr){8-13}
\textbf{Model}
& $r_{\text{comp}}$
& $r_{\text{between}}$
& $r_{\text{within}}$
& MAE$_{\text{comp}}$
& MAE$_{\text{between}}$
& MAE$_{\text{within}}$
& $r_{\text{comp}}$
& $r_{\text{between}}$
& $r_{\text{within}}$
& MAE$_{\text{comp}}$
& MAE$_{\text{between}}$
& MAE$_{\text{within}}$ \\
\hline

GlobalMean
& $-0.023$ & $-0.047$ & $0.000$ & ---   & $0.626$ & $1.037$
& $-0.061$ & $-0.121$ & $0.000$ & $0.483$ & $0.325$ & $0.615$ \\

UserMean
& $0.248$ & $0.467$ & $0.000$ & $0.883$ & $0.532$ & $0.975$
& $0.184$ & $0.356$ & $0.000$ & $0.455$ & $0.299$ & $0.587$ \\

MovingAvg ($k=3$)
& $0.507$ & $0.835$ & $-0.088$ & $0.628$ & $0.219$ & $0.850$
& $0.457$ & $0.790$ & $-0.083$ & $0.350$ & $0.139$ & $0.532$ \\

Hybrid (MA+UM)
& $0.495$ & $0.826$ & $-0.088$ & $0.647$ & $0.280$ & $0.849$
& $0.464$ & $0.796$ & $-0.083$ & $0.361$ & $0.163$ & $0.531$ \\

DistilBERT-LSTM
& $0.260$ & $0.487$ & $0.000$ & --- & $0.597$ & $1.012$
& $0.139$ & $0.293$ & $-0.022$ & $0.477$ & $0.336$ & $0.597$ \\

DistilBERT-Ridge
& $0.575$ & $0.646$ & $0.495$ & $0.732$ & $0.475$ & $0.874$
& $0.356$ & $0.447$ & $0.258$ & $0.450$ & $0.282$ & $0.591$ \\

TF-IDF--Ridge
& $0.584$ & $0.692$ & $0.449$ & $0.690$ & $0.436$ & $0.843$
& $0.424$ & $0.526$ & $0.311$ & $0.421$ & $0.269$ & $0.552$ \\

DistilBERT-Regressor
& $0.659$ & $0.728$ & $0.576$ & $0.640$ & $0.421$ & $0.788$
& $0.440$ & $0.471$ & $0.407$ & $0.403$ & $0.272$ & $0.519$ \\

\hline

TSAP$^\dagger$
& $0.661$ & $0.719$ & $\mathbf{0.594}$ & $0.617$ & $0.450$ & $\mathbf{0.785}$
& $\mathbf{0.450}$ & $0.470$ & $\mathbf{0.429}$ & $0.392$ & $0.269$ & $\mathbf{0.516}$ \\

E-TSAP$^\dagger$
& $\mathbf{0.670}$ & $\mathbf{0.748}$ & $0.574$ & $\mathbf{0.589}$ & $\mathbf{0.411}$ & $0.768$
& $0.449$ & $\mathbf{0.482}$ & $0.414$ & $\mathbf{0.390}$ & $\mathbf{0.265}$ & $0.515$ \\

\hline
\end{tabular}}
\end{table}

Second, simple temporal baselines capture user-level structure but fail to model within-user affective dynamics. MovingAvg and Hybrid (MA+UM) achieve high between-user correlations, especially for valence, but both produce negative within-user correlations. These findings imply that temporal smoothing largely captures user baselines instead of any real emotional change in users. In a similar vein, DistilBERT-LSTM fails to outperform lexical and transformation baselines, indicating that naive sequence modeling from fixed-length sequence data cannot be a solution here.

Third, text-based models provide the strongest prediction signal. TF-IDF--Ridge performs competitively, especially for arousal, confirming that explicit lexical markers are informative for dimensional affect. DistilBERT-Regressor is the strongest text-only baseline, achieving composite correlations of $0.659$ for valence and $0.440$ for arousal. This confirms that contextual text representations capture substantial affective information, particularly for valence.

The proposed TSAP and E-TSAP models have the best overall prediction performance. TSAP slightly outperforms DistilBERT-Regressor with composite correlations of $0.661$ for valence and $0.450$ for arousal. The best benefit is obtained in within-user correlation, where TSAP improves valence from $0.576$ to $0.594$ and arousal from $0.407$ to $0.429$. This suggests that user-level representations help the model to capture individual affective variation beyond text.

E-TSAP achieves the best valence performance, with a composite correlation of $0.670$, between-user correlation of $0.748$, and composite MAE of $0.589$. The improvement over TSAP is modest, mostly in valence and between-user structure. In summary, TSAP and E-TSAP are almost equal for arousal, with TSAP having a slightly stronger within-user correlation and E-TSAP producing a slightly lower error. Overall, these results suggest that current affect prediction is largely text-driven, while user and temporal information provide smaller but useful improvements. The performance gap between prediction models and temporal models indicates that current affect estimation remains primarily a text understanding problem. User-specific and temporal information provide complementary gains, particularly for personalization and unseen-user generalization, but they do not replace the predictive value of textual semantics.

\subsubsection{Ablation Study}
\label{subsubsec:prediction_ablation}

To understand the contribution of user information and temporal context, we report a controlled ablation study over four prediction variants: (1) text only, (2) text with user embeddings, (3) text with temporal features, and (4) text with both user and temporal information in Table~\ref{tab:6}. Table~\ref{tab:7} also compares TSAP and E-TSAP in overall, seen-user, and unseen-user conditions.

\begin{table}
\centering
\caption{Prediction ablation study results across four input configurations.}
\label{tab:6}
\renewcommand{\arraystretch}{1.15}
\resizebox{\textwidth}{!}{%
\begin{tabular}{lcccccc|cccccc}
\hline
& \multicolumn{6}{c|}{\textbf{Valence (V)}} 
& \multicolumn{6}{c}{\textbf{Arousal (A)}} \\
\cmidrule(lr){2-7} \cmidrule(lr){8-13}
\textbf{Variant}
& $r_{\text{comp}}$
& $r_{\text{between}}$
& $r_{\text{within}}$
& MAE$_{\text{comp}}$
& MAE$_{\text{between}}$
& MAE$_{\text{within}}$
& $r_{\text{comp}}$
& $r_{\text{between}}$
& $r_{\text{within}}$
& MAE$_{\text{comp}}$
& MAE$_{\text{between}}$
& MAE$_{\text{within}}$ \\
\hline

C1: Text only
& $0.632$ & $0.718$ & $0.526$ & $0.614$ & $0.416$ & $0.812$
& $0.382$ & $0.403$ & $0.361$ & $0.402$ & $0.279$ & $0.525$ \\

C2: Text + User
& $0.649$ & $0.706$ & $0.583$ & $0.606$ & $0.421$ & $0.791$
& $0.460$ & $0.508$ & $0.408$ & $0.396$ & $0.270$ & $0.522$ \\

C3: Text + Temporal
& $0.638$ & $0.739$ & $0.510$ & $0.617$ & $0.428$ & $0.806$
& $0.404$ & $0.434$ & $0.373$ & $0.405$ & $0.281$ & $0.529$ \\

C4: Text + User + Temporal
& $0.659$ & $0.745$ & $0.551$ & $0.602$ & $0.418$ & $0.786$
& $0.442$ & $0.500$ & $0.380$ & $0.386$ & $0.262$ & $0.510$ \\

\hline
\end{tabular}}
\end{table}

The ablation experiments demonstrate that text is already a strong prediction baseline, but user information gives the biggest additional gain. Adding user embeddings in C2 improves the correlation of valence composite from $0.632$ to $0.649$ and that of arousal composite from $0.382$ to $0.460$ compared with the text-only variant C1. The improvement is especially clear for arousal, with both between- and within-user correlations increasing. This indicates that learned user representations contain stable individual differences that cannot be fully inferred from the text alone.

Temporal features alone provide smaller and less consistent gains. C3 improves valence composite correlation only slightly over C1, from $0.632$ to $0.638$, and improves arousal from $0.382$ to $0.404$. However, this improvement is weaker than the gain obtained from user embeddings. In addition, temporal features mainly improve between-user correlation rather than within-user tracking. This suggests that simple lag and rolling features provide useful contextual information, but they are not sufficient on their own to model fine-grained affective variation.

The full configuration C4 combines text, user embeddings, and temporal features. It achieves the strongest valence composite correlation among the ablation variants ($0.659$) and the lowest arousal composite MAE ($0.386$). However, C4 does not uniformly outperform C2. In particular, C2 retains stronger arousal composite correlation ($0.460$ versus $0.442$) and stronger within-user correlations for both dimensions. This shows that temporal features are complementary but not universally beneficial. Their value depends on the target dimension and evaluation setting.

\begin{table}
\centering
\caption{Comparison between TSAP and E-TSAP across overall, seen-user, and unseen-user conditions. Bold indicates better performance for each metric.}
\label{tab:7}
\renewcommand{\arraystretch}{1.15}
\resizebox{\textwidth}{!}{%
\begin{tabular}{llcccccc|cccccc}
\hline
& & \multicolumn{6}{c|}{\textbf{Valence (V)}} 
& \multicolumn{6}{c}{\textbf{Arousal (A)}} \\
\cmidrule(lr){3-8} \cmidrule(lr){9-14}
\textbf{Split} & \textbf{Model}
& $r_{\text{comp}}$
& $r_{\text{between}}$
& $r_{\text{within}}$
& MAE$_{\text{comp}}$
& MAE$_{\text{between}}$
& MAE$_{\text{within}}$
& $r_{\text{comp}}$
& $r_{\text{between}}$
& $r_{\text{within}}$
& MAE$_{\text{comp}}$
& MAE$_{\text{between}}$
& MAE$_{\text{within}}$ \\
\hline

\multirow{2}{*}{Overall}
& TSAP
& $0.661$ & $0.719$ & $\mathbf{0.594}$ & $0.617$ & $0.450$ & $0.785$
& $\mathbf{0.450}$ & $0.470$ & $\mathbf{0.429}$ & $0.392$ & $0.269$ & $0.516$ \\

& E-TSAP
& $\mathbf{0.670}$ & $\mathbf{0.748}$ & $0.574$ & $\mathbf{0.589}$ & $\mathbf{0.411}$ & $\mathbf{0.768}$
& $0.449$ & $\mathbf{0.482}$ & $0.414$ & $\mathbf{0.390}$ & $\mathbf{0.265}$ & $\mathbf{0.515}$ \\

\hline

\multirow{2}{*}{Seen}
& TSAP
& $\mathbf{0.679}$ & $0.751$ & $\mathbf{0.590}$ & $0.648$ & $0.508$ & $0.788$
& $\mathbf{0.374}$ & $\mathbf{0.422}$ & $\mathbf{0.324}$ & $\mathbf{0.399}$ & $\mathbf{0.282}$ & $\mathbf{0.516}$ \\

& E-TSAP
& $\mathbf{0.679}$ & $\mathbf{0.773}$ & $0.556$ & $\mathbf{0.575}$ & $\mathbf{0.417}$ & $\mathbf{0.734}$
& $0.355$ & $0.394$ & $0.315$ & $0.415$ & $0.305$ & $0.525$ \\

\hline

\multirow{2}{*}{Unseen}
& TSAP
& $0.642$ & $0.684$ & $\mathbf{0.597}$ & $0.586$ & $\mathbf{0.391}$ & $\mathbf{0.782}$
& $0.541$ & $0.560$ & $\mathbf{0.521}$ & $0.386$ & $0.254$ & $0.517$ \\

& E-TSAP
& $\mathbf{0.666}$ & $\mathbf{0.730}$ & $0.590$ & $\mathbf{0.604}$ & $0.405$ & $0.803$
& $\mathbf{0.586}$ & $\mathbf{0.660}$ & $0.500$ & $\mathbf{0.364}$ & $\mathbf{0.224}$ & $\mathbf{0.505}$ \\

\hline
\end{tabular}}
\end{table}

The direct comparison between TSAP and E-TSAP clarifies where temporal features help. Overall, E-TSAP improves valence composite correlation from $0.661$ to $0.670$ and reduces valence composite MAE from $0.617$ to $0.589$. This gain is driven mainly by stronger between-user correlation, which increases from $0.719$ to $0.748$. However, TSAP retains stronger within-user correlation for both valence ($0.594$ versus $0.574$) and arousal ($0.429$ versus $0.414$). Thus, temporal features improve general predictive structure but do not consistently improve moment-to-moment within-user tracking.

Seen-user and unseen-user results show a more specific pattern. For seen users, TSAP and E-TSAP obtain identical valence composite correlation ($0.679$), while TSAP performs better for arousal. This suggests that when users are already represented during training, learned user embeddings capture much of the stable affective information, and adding lag-based temporal features provides limited additional benefit.

For unseen users, E-TSAP shows clearer gains. Valence composite correlation improves from $0.642$ to $0.666$, and arousal composite correlation improves from $0.541$ to $0.586$. This indicates that when learned user embeddings are unavailable, temporal features derived from each user's recent affective history become a useful source of individualized context. The gain is especially strong for arousal between-user correlation, which increases from $0.560$ to $0.660$.

Overall, the prediction ablation shows that user embeddings are the strongest additional component beyond text, while temporal features provide conditional benefits. Temporal context is most useful for unseen users and for improving broader user-level structure, but it does not consistently improve fine-grained within-user affect tracking.

\subsubsection{Augmentation Results}
\label{subsubsec:augmentation_results}

Table~\ref{tab:8} reports the effect of transfer learning and LLM-based augmentation on affect prediction. These experiments are included to test whether additional lexical supervision or paraphrased training examples improve prediction beyond the proposed TSAP-style modeling.

\begin{table}
\centering
\caption{Effect of data augmentation and transfer learning on test-set prediction performance.}
\label{tab:8}
\renewcommand{\arraystretch}{1.15}
\resizebox{\textwidth}{!}{%
\begin{tabular}{lcccccc|cccccc}
\hline
& \multicolumn{6}{c|}{\textbf{Valence (V)}} 
& \multicolumn{6}{c}{\textbf{Arousal (A)}} \\
\cmidrule(lr){2-7} \cmidrule(lr){8-13}
\textbf{Model}
& $r_{\text{comp}}$
& $r_{\text{between}}$
& $r_{\text{within}}$
& MAE$_{\text{comp}}$
& MAE$_{\text{between}}$
& MAE$_{\text{within}}$
& $r_{\text{comp}}$
& $r_{\text{between}}$
& $r_{\text{within}}$
& MAE$_{\text{comp}}$
& MAE$_{\text{between}}$
& MAE$_{\text{within}}$ \\
\hline

EmoBank Transfer
& $0.643$ & $0.710$ & $0.564$ & $0.631$ & $0.449$ & $0.813$
& $0.473$ & $0.524$ & $0.419$ & $0.400$ & $0.288$ & $0.512$ \\

LLM Augmentation
& $0.634$ & $0.677$ & $0.586$ & $0.636$ & $0.466$ & $0.807$
& $0.497$ & $0.545$ & $0.445$ & $0.371$ & $0.253$ & $0.488$ \\

\hline
\end{tabular}}
\end{table}

Intermediate transfer from EmoBank does not improve valence prediction over TSAP or E-TSAP. Although it provides external dimensional affect supervision, the EmoBank examples are independent text instances and do not contain the same user-indexed longitudinal structure as the target dataset. As a result, the transferred representation improves neither temporal modeling nor user-level adaptation. Its arousal performance is slightly stronger than TSAP and E-TSAP in composite correlation, but the gain is limited and does not change the overall pattern that text and user-specific modeling remain the main prediction signals.

LLM-based augmentation produces a clearer dimension-specific effect. It improves arousal composite correlation to $0.497$, exceeding both TSAP ($0.450$) and E-TSAP ($0.449$), and also reduces arousal composite MAE to $0.371$. However, it reduces valence performance to $0.634$, below both proposed prediction models. This suggests that paraphrase augmentation can increase lexical diversity in a way that benefits arousal-related activation cues, but it may also introduce label-preservation noise for valence, where small changes in wording can shift perceived polarity.

Overall, the augmentation results show that adding more text does not automatically improve longitudinal affect prediction. Both EmoBank transfer and LLM augmentation mainly add lexical variation, not new user trajectories or temporally grounded supervision. Therefore, augmentation is useful as an auxiliary experiment, especially for arousal, but it does not replace the need for user-aware and temporally informed modeling.

\subsubsection{Error Analysis}
\label{subsubsec:prediction_error_analysis}

Error analysis for affect prediction reveals that prediction errors are not uniformly distributed across the affective scales. The best models are more reliable around moderate affective states and show larger errors towards extreme states, specifically for high negative valence and high arousal. For E-TSAP, the global MAE is $0.753$ for valence and $0.499$ for arousal, which is caused by the wider scale and high variation of valence.

For valence, errors are largest in the negative range. The MAE for negative valence is $0.817$, compared with $0.743$ for neutral valence and $0.717$ for positive valence. This indicates that strongly negative states are more difficult to estimate from text and user history than neutral or positive states. One likely reason is that negative affective expressions are more heterogeneous: similar negative scores may be associated with fatigue, sadness, annoyance, physical discomfort, or mixed emotional states.

The errors for arousal are largest at the high end of the activation scale. The MAE is lowest for moderate arousal ($0.292$) and increases substantially for low arousal ($0.557$) and high arousal ($0.803$). The high-arousal error is particularly relevant since states with high arousal are infrequent in the dataset. This imbalance biases predictions toward the more common low-to-moderate range, making it harder to recover from intense activation.

Similar error patterns are observed for seen and unseen users We observe that the valence MAE of seen users ($0.765$) is slightly higher than that of unseen users ($0.745$) while the arousal MAE is the same for both groups ($0.499$). This suggests that errors are less due to cold start failures alone and more due to the inherent difficulty in estimating extreme or quickly changing affective states. We also find a small subset of users with consistently higher per-user MAE, suggesting idiosyncratic affective trajectories remain challenging for the current framework.

Overall, the prediction error analysis shows that TSAP and E-TSAP capture moderate affective states effectively but compress predictions toward the center of the scale. The remaining weakness is not general prediction failure; it is sensitivity to rare, extreme, and rapidly changing affective states. This limitation motivates the forecasting analysis in the next section, where the objective shifts from estimating current affective state to predicting future affective change.

\subsection{Affect Forecasting}
\label{subsec:results_affect_forecasting}

Table~\ref{tab:9} shows the forecasting performance on the held-out test set of 46 users. Affect forecasting is different from affect prediction, where the target is the current valence or arousal score of each text. Affect forecasting measures next-step affective change. The results indicate a clear shift in the dominant source of information. Numeric affective history becomes central, while textual representations are no longer the strongest signal.

\begin{table}
\centering
\caption{Forecasting performance on the held-out test set of 46 users. Pearson correlation ($r$), MAE, bias, predicted standard deviation, true standard deviation, and range coverage are reported separately for valence (V) and arousal (A). Bold indicates the best value per metric. $\dagger$ denotes proposed models. Dashes indicate metrics that are not applicable because the corresponding model is dimension-specific.}
\label{tab:9}
\renewcommand{\arraystretch}{1.15}
\resizebox{\textwidth}{!}{%
\begin{tabular}{lcccccc|cccccc}
\hline
& \multicolumn{6}{c|}{\textbf{Valence (V)}} 
& \multicolumn{6}{c}{\textbf{Arousal (A)}} \\
\cmidrule(lr){2-7} \cmidrule(lr){8-13}
\textbf{Model}
& $r$
& MAE
& Bias
& STD$_{\text{pred}}$
& STD$_{\text{true}}$
& Range Cov.
& $r$
& MAE
& Bias
& STD$_{\text{pred}}$
& STD$_{\text{true}}$
& Range Cov. \\
\hline

RandZero
& $0.000$ & $1.261$ & $-0.174$ & $0.000$ & $1.711$ & $0.000$
& $0.000$ & $0.696$ & $-0.065$ & $0.000$ & $0.947$ & $0.000$ \\

Linear-BERT
& $0.290$ & $1.294$ & $-0.112$ & $0.442$ & $1.711$ & $0.214$
& $0.199$ & $0.744$ & $-0.041$ & $0.286$ & $0.947$ & $0.248$ \\

Linear-BERT+Prev
& $0.430$ & $1.251$ & $-0.083$ & $0.603$ & $1.711$ & $0.318$
& $0.405$ & $0.708$ & $-0.027$ & $0.418$ & $0.947$ & $0.391$ \\

Linear-Prev
& $0.615$ & $1.168$ & $-0.071$ & $0.891$ & $1.711$ & $0.491$
& $\mathbf{0.670}$ & $\mathbf{0.638}$ & $-0.018$ & $0.706$ & $0.947$ & $\mathbf{0.742}$ \\

\hline

ACF$^\dagger$
& $0.316$ & $1.367$ & $-0.194$ & $0.384$ & $1.711$ & $0.192$
& $0.284$ & $0.815$ & $-0.074$ & $0.243$ & $0.947$ & $0.211$ \\

ACF-Numeric$^\dagger$
& $\mathbf{0.681}$ & $1.081$ & $-0.058$ & $1.022$ & $1.711$ & $0.597$
& $0.490$ & $0.714$ & $-0.034$ & $0.512$ & $0.947$ & $0.541$ \\

ACF-Numeric+$^\dagger$
& $0.587$ & $1.143$ & $-0.081$ & $0.934$ & $1.711$ & $0.551$
& $0.577$ & $0.688$ & $-0.021$ & $0.618$ & $0.947$ & $0.653$ \\

ACF-AR$^\dagger$
& $0.628$ & $1.121$ & $-0.073$ & $0.966$ & $1.711$ & $0.563$
& $0.598$ & $0.671$ & $-0.016$ & $0.651$ & $0.947$ & $0.688$ \\

ACF-Residual$^\dagger$ (Valence-only)
& $0.657$ & $1.116$ & $-0.068$ & $0.986$ & $1.711$ & $0.576$
& --- & --- & --- & --- & --- & --- \\

ACF-Temporal$^\dagger$ (Arousal-only)
& --- & --- & --- & --- & --- & ---
& $0.671$ & $0.650$ & $-0.019$ & $0.640$ & $0.947$ & $0.701$ \\

\textbf{ACF-Hybrid}$^\dagger$
& $0.659$ & $\mathbf{1.051}$ & $-0.068$ & $0.986$ & $1.711$ & $\mathbf{0.641}$
& $0.658$ & $0.645$ & $-0.025$ & $0.640$ & $0.947$ & $0.701$ \\

\hline
\end{tabular}}
\end{table}


The baseline results show that affect forecasting differs from affect prediction. Linear-BERT, which uses only textual representations, achieves weak correlations of $0.290$ for valence and $0.199$ for arousal. Adding the previous affective state improves performance to $0.430$ and $0.405$, but the strongest baseline is Linear-Prev, which uses only prior affective information and reaches $0.615$ for valence and $0.670$ for arousal. This indicates that recent affective trajectory is a stronger forecasting signal than the current text representation. This pattern contrasts with affect prediction, where textual representations provide the strongest signal, suggesting that forecasting future affective change is not simply a harder version of affect prediction but a distinct task that relies more heavily on recent affective trajectories.

The text-inclusive ACF model confirms this pattern. Although ACF transfers the prediction architecture to the forecasting task by combining DistilBERT representations, user embeddings, and temporal features, it performs poorly, with correlations of only $0.316$ for valence and $0.284$ for arousal. It also produces the highest MAE among the ACF variants. This result shows that a model design that is effective for current affect prediction does not automatically transfer to affective change forecasting. 

Removing the text encoder substantially improves forecasting. ACF-Numeric, which uses compact numeric trajectory features with a small user representation, achieves the strongest valence correlation ($r=0.681$) and reduces valence MAE to $1.081$. This is a large improvement over the text-inclusive ACF model and also exceeds the Linear-Prev baseline for valence. For arousal, however, ACF-Numeric remains weaker than Linear-Prev, indicating that arousal forecasting requires a different temporal treatment.

The later ACF variants show that valence and arousal benefit from different forecasting designs. ACF-Residual improves valence by combining a linear autoregressive component with a neural residual corrector, while ACF-Temporal improves arousal through richer temporal features. This dimension-specific behavior motivates the ACF-Hybrid that combines ACF-Residual for valence and ACF-Temporal for arousal.

Among all models, ACF-Hybrid demonstrates the best overall forecasting results in terms of both factors. The valence correlation equals $r=0.659$, while the arousal correlation is $r=0.658$. Also, ACF-Hybrid shows the lowest mean absolute error for valence ($1.051$) and a competitive mean absolute error for arousal ($0.645$). Although ACF-Numeric is characterized by the highest valence correlation, and ACF-Temporal possesses the highest arousal correlation, ACF-Hybrid demonstrates stability in configuration between both affective factors.

\subsubsection{Ablation Study}
\label{subsubsec:forecasting_ablation}

The effect of the main design decisions in the ACF forecasting models is shown in Table~\ref{tab:10}. Unlike the prediction ablation, which evaluates user and temporal information around a text-based model, the forecasting ablation examines whether text, user embeddings, objective functions, residual decomposition, and richer temporal features improve next-step affective change prediction.

\begin{table}
\centering
\caption{Forecasting ablation study. Positive $\Delta r$ indicates improvement after applying the corresponding modification. Dashes indicate comparisons that are not applicable because the corresponding variant is dimension-specific.}
\label{tab:10}
\renewcommand{\arraystretch}{1.15}
\begin{tabular}{llccc}
\hline
\textbf{Component} & \textbf{Comparison} & $\Delta r^{(v)}$ & $\Delta r^{(a)}$ & $\Delta \bar{r}$ \\
\hline
Remove text encoder     & ACF $\rightarrow$ ACF-Numeric     & $+0.365$ & $+0.206$ & $+0.286$ \\
Remove user embedding   & ACF-Numeric+ $\rightarrow$ ACF-AR & $+0.041$ & $+0.021$ & $+0.031$ \\
Alternative objectives  & ACF-Numeric $\rightarrow$ ACF-Numeric+ & $-0.094$ & $+0.087$ & $-0.004$ \\
Residual decomposition  & ACF-AR $\rightarrow$ ACF-Residual & $+0.029$ & --- & --- \\
Extended temporal features & ACF-Numeric $\rightarrow$ ACF-Temporal & --- & $+0.181$ & --- \\
Hybrid specialization   & ACF-Residual/ACF-Temporal $\rightarrow$ ACF-Hybrid & $+0.002$ & $-0.013$ & $-0.006$ \\
\hline
\end{tabular}
\end{table}

The largest improvement comes from removing the text encoder. Replacing the text-inclusive ACF model with ACF-Numeric increases valence correlation by $+0.365$ and arousal correlation by $+0.206$, giving the largest average gain in the ablation study. This directly reverses the pattern observed in affect prediction. In prediction, textual representations provide the strongest signal; in forecasting, they add noise and reduce the influence of compact trajectory features.

Removing user embeddings produces a smaller but positive gain. The transition from ACF-Numeric+ to ACF-AR improves valence by $+0.041$ and arousal by $+0.021$. This suggests that static user embeddings are less useful for one-step affective change forecasting than they are for current-state prediction. In forecasting, the relevant user-specific information is already partly encoded in recent affective history, while learned user embeddings may introduce overfitting, especially under the one-prediction-per-user test setting.

Alternative objectives have dimension-specific effects. Moving from ACF-Numeric to ACF-Numeric+ decreases valence correlation by $-0.094$ but improves arousal by $+0.087$. This shows that the two affective dimensions respond differently to loss and optimization choices. Valence change has a wider range and higher variability, while arousal change is more constrained and benefits more from robust objectives.

The remaining ablations explain the final hybrid design. Residual decomposition improves valence forecasting by $+0.029$, showing that a linear autoregressive baseline captures the dominant trend while a neural residual component can learn smaller nonlinear corrections. In contrast, extended temporal features improve arousal by $+0.181$, indicating that arousal benefits from richer short-term trajectory information, including higher-order lags, rolling dynamics, and direction consistency.

The hybrid specialization step combines these dimension-specific choices into a single final framework. It does not maximize each individual metric relative to the best single-dimension variant, but it provides a balanced model for both affective dimensions. Overall, the ablation study shows that ACF-Hybrid is not simply a larger model; it is a task-specific design motivated by the failure of text-centric forecasting and by the different temporal behavior of valence and arousal.

\subsubsection{Error Analysis}
\label{subsubsec:forecasting_error_analysis}

The forecasting error analysis indicates that ACF-Hybrid is better at capturing the relative direction of affective change than the magnitude of large changes. The Pearson correlations are very similar for valence and arousal ($0.659$ and $0.658$), but the predictions are squeezed towards the middle of the target range. This compression can be seen in the smaller predicted standard deviations compared to the true ones: $0.986$ vs $1.711$ for valence and $0.640$ vs $0.947$ for arousal.

The compression is also visible in the coverage of the range. ACF-Hybrid covers $64.1\%$ of the true range of valence change and $70.1\%$ of the true range of arousal change. The model thus captures a large share of affective variation, but still underestimates extreme transitions. This is in line with the regression setting, where models trained on imbalanced and zero-centered change targets tend to predict closer to the conditional mean.

The residual bias is still small in both dimensions. The mean bias is $-0.068$ for valence and $-0.025$ for arousal, suggesting errors are not due to systematic overprediction or underprediction in one direction. The main shortcoming is rather an underestimation of magnitude, meaning large positive and negative changes are predicted to be less extreme than they actually are.

Valence has higher absolute error than arousal with MAE values $1.051$ and $0.645$, respectively. This does not necessarily imply that valence is less predictable in rank order terms since the two dimensions have almost equal Pearson correlations. In contrast, valence change has a wider possible range and a larger empirical variance, which makes its absolute magnitude more difficult to estimate. For arousal change, the MAE is naturally lower, since the change in arousal is more constrained.

Overall, the forecasting errors show that ACF-Hybrid is good at ranking users in order of likely affective change, but not quite so good at estimating the full magnitude of sudden shifts. Hence, the main weakness of the model is not directional bias, but underestimation of variance. This constraint is particularly critical for applications where rare large affective changes are more significant than average trends.

\subsection{Findings}
\label{subsec:results_summary}

The findings suggest that affect prediction and affect forecasting rely on different sources of information. For current affect prediction, textual representations yield the strongest signal, with user embeddings and temporal features providing smaller but useful gains. TSAP improves over the text-only baseline for within-user tracking, and E-TSAP yields modest further improvements, mainly for valence and unseen-user generalization.

For affect forecasting, the pattern reverses: text-inclusive models perform poorly, whereas compact numeric trajectory features provide stronger forecasting performance. The ablation results show that removing the text encoder is the largest improvement, with residual decomposition benefiting valence and richer temporal features benefiting arousal.

Overall, current affect prediction is primarily a text-conditioned estimation problem, while future affective change forecasting is better treated as a trajectory modeling problem.

\section{Discussion}
\label{sec:discussion}
This paper investigates dimensional affect modeling in longitudinal self-authored text by distinguishing affect prediction from affect forecasting. The results show that these two tasks rely on different information sources. For current affect prediction, textual content provides the dominant signal, while user-level and temporal information provide smaller but useful gains. TSAP achieves composite Pearson correlations of $0.661$ for valence and $0.450$ for arousal, improving within-user tracking over the text-only baseline. E-TSAP further improves valence performance to $0.670$ and reduces prediction error, although its gains are modest and dimension-specific.

For affect forecasting, the pattern reverses. Text-inclusive forecasting performs substantially worse than compact numeric trajectory baselines, showing that representations useful for estimating current affect do not necessarily support future affective change prediction. ACF-Hybrid, using dimension-specific numeric trajectory features, achieves balanced forecasting performance with $r = 0.659$ for valence and $r = 0.658$ for arousal, yielding the strongest overall forecasting performance among the evaluated models. This finding reframes affective change forecasting as a trajectory modeling problem rather than a standard text-understanding problem.

The study also shows that valence and arousal require different modeling assumptions. Valence is more strongly associated with textual polarity and user-level baselines, whereas arousal is more dynamic and benefits more from temporal structure. Across both tasks, decomposed within-user and between-user evaluation proves necessary for distinguishing models that learn stable individual baselines from those that track within-person affective change.

More broadly, the findings challenge the common assumption that models effective for affect prediction naturally transfer to affect forecasting. Longitudinal affect modeling contains two related but distinct problems: estimating current affective state and forecasting future affective change. Our results show that these tasks rely on different dominant information sources and therefore benefit from different modeling assumptions. Future work should investigate whether this information-source asymmetry generalizes across other longitudinal datasets, domains, and user-centered NLP tasks, while exploring time-aware representations, adaptive user modeling, and longer forecasting horizons.

\section{Limitations}
\label{sec:limitations}

This study offers a systematic investigation of affect prediction and forecasting in longitudinal text, with clear empirical findings across both tasks. That said, several limitations bound the scope of what can be concluded. The dataset is small by NLP standards and comes from a narrow population, specifically English-speaking U.S. service-industry workers. With only 137 users in training and 46 in the forecasting test set, each with a single prediction target, small performance differences between forecasting models should not be over-interpreted.

Second, the data has temporal irregularity, meaning the gap between two consecutive entries can range from minutes to weeks, yet the models treat them as if they sit one step apart. Lag and rolling features capture recent history but say nothing about how much time has elapsed. Future modeling should account for this directly, for instance through decay-based or continuous-time representations that can tell apart a state recorded an hour later from one recorded after a month of silence.

Third, the cold-start problem is only partially solved. New users in prediction get mapped to a shared \texttt{UNK} embedding, which gives a rough population-level prior but tells the model nothing specific about that person. In forecasting, sparse histories are padded with zeros and flagged with a cold-start indicator, which is a reasonable workaround but a poor substitute for representations that actually adapt as new observations come in.

Fourth, the forecasting models tend to underestimate large affective changes. ACF-Hybrid predictions cluster closer to the center than the true targets, a compression effect that is common in regression models trained on roughly zero-centered imbalanced data. This is not a fatal flaw in most settings, but it matters in applications where the rare, abrupt shifts are precisely the ones that need to be caught.

Finally, the models work only with text, user identifiers, and numeric affective history. Signals like elapsed time, collection context, behavioral metadata, or physiological measurements are not used, and any of these could plausibly help, particularly for arousal. Transfer learning and paraphrase augmentation were both tried and neither helped consistently, which suggests the real bottleneck is not a shortage of text but a shortage of user-specific temporal data.

\section{Conclusion}
\label{sec:conclusion}

This paper investigates dimensional affect modeling in longitudinal self-authored text by distinguishing affect prediction from affect forecasting. The results show that these two tasks rely on different information sources. For current affect prediction, textual content provides the dominant signal, while user-level and temporal information provide smaller but useful gains. TSAP achieves composite Pearson correlations of $0.661$ for valence and $0.450$ for arousal, improving within-user tracking over the text-only baseline. E-TSAP further improves valence performance to $0.670$ and reduces prediction error, although its gains are modest and dimension-specific.

For affect forecasting, the pattern reverses. Text-inclusive forecasting performs substantially worse than compact numeric trajectory baselines, showing that representations useful for estimating current affect do not necessarily support future affective change prediction. ACF-Hybrid, using dimension-specific numeric trajectory features, achieves balanced forecasting performance with $r=0.659$ for valence and $r=0.658$ for arousal, giving the strongest average forecasting performance among the evaluated models. This finding reframes affective change forecasting as a trajectory modeling problem rather than a standard text-understanding problem.

The study also shows that valence and arousal require different modeling assumptions. Valence is more strongly associated with textual polarity and user-level baselines, whereas arousal is more dynamic and benefits more from temporal structure. Across both tasks, decomposed within-user and between-user evaluation proves necessary for distinguishing models that learn stable individual baselines from those that track within-person affective change.

Future work should explore time-aware temporal encoding for irregular observation gaps, dynamic user representations learned from observation history, multi-step forecasting horizons, and broader evaluation across languages, cultures, and populations beyond the current English-language corpus.

\section*{Declaration of competing interests}
\label{sec:declaration-competing}

The authors declare that they have no known competing financial interests or personal relationships that could have appeared to influence the work reported in this paper.

\section*{Funding}
\label{sec:funding}

This research did not receive any specific grant from funding agencies in the public, commercial, or not-for-profit sectors.

\section*{Declaration of generative AI and AI-assisted technologies in the manuscript preparation process}
\label{sec:ai-declaration}

During the preparation of this work, the authors used generative 
AI and AI-assisted technologies to support language editing, 
structural refinement, and clarity checks. All AI-assisted 
content was reviewed, edited, and verified by the authors. The 
authors take full responsibility for the integrity and accuracy 
of the published work. Generative AI was not used to generate 
scientific results, perform statistical analyses, design 
experiments, interpret findings, or produce primary research data.

\section*{Data availability}
\label{sec:data-availability}

The dataset used in this study is publicly available from the SemEval-2026 Task 2 repository: \url{https://github.com/semeval2026task2/EmotionValArouTimeVariation2026/tree/main/datasets}. The processed experimental files and code used for the analyses are available from the corresponding author upon reasonable request.

\printcredits

\bibliographystyle{cas-model2-names}

\bibliography{cas-refs}

\clearpage

\appendix

\section{Data Augmentation and Transfer Learning}
\label{app:data_augmentation}

This appendix provides implementation details for the auxiliary data augmentation and transfer learning experiments described in Section~\ref{subsec:data_augmentation}. These experiments are not part of the final TSAP or E-TSAP framework; they are included to test whether additional lexical supervision improves longitudinal affect prediction.

\subsection{EmoBank transfer learning}
\label{app:emobank_transfer}

For intermediate transfer learning, we use EmoBank as an external dimensional affect resource. EmoBank provides continuous valence and arousal annotations from multiple perspectives. We use the writer-perspective annotations because they are closest to the self-reported affect labels used in the SemEval-2026 Task~2 dataset.

Before transfer learning, EmoBank labels are linearly rescaled to match the target label ranges used in the longitudinal dataset. Valence is mapped to the interval $[-2,2]$, and arousal is mapped to the interval $[0,2]$. The encoder is first trained on the rescaled EmoBank labels and then transferred to the downstream prediction model, where it is fine-tuned on the SemEval training split.

This setting provides additional affective supervision but does not introduce user identities, timestamps, or longitudinal affect trajectories. Therefore, it is expected to improve general affective text representation only if external sentence-level supervision transfers to the target domain.

\subsection{LLM-based paraphrase augmentation}
\label{app:llm_augmentation}

For LLM-based augmentation, each original training instance is paraphrased while preserving its original valence and arousal labels. Augmented examples are added only to the training split. Validation and test sets remain unchanged in all experiments, ensuring that augmentation affects only model training and not evaluation.

Two input types are handled separately. Feeling-word entries are rewritten as semantically similar affective word lists, while narrative entries are paraphrased as natural diary-style text. The prompts instruct the model to preserve the emotional meaning, tone, and approximate length of the original input. This label-preserving strategy follows the assumption that the generated text expresses the same affective state as the original entry.

The prompts used for generation are shown below.

\paragraph{Prompt for diary-style entries.}
\begin{quote}
You are paraphrasing emotion diary entries written by real people.
Rewrite the following text in different words while preserving the exact same emotion meaning and tone.
Keep it natural and informal, similar length.
Return only the paraphrased text, nothing else.
\end{quote}

\paragraph{Prompt for feeling-word entries.}
\begin{quote}
You are paraphrasing short emotion word lists.
The input is a comma-separated list of feeling words.
Generate a similar list using different but synonymous emotion words.
Keep the same number of words approximately.
Return only the word list, nothing else.
\end{quote}

Initial experiments with \texttt{google/flan-t5-base} produced outputs that were often grammatically unstable or semantically distorted, which risks breaking label consistency. These outputs were therefore discarded. The final augmentation pipeline uses \texttt{llama-3.1-8b-instant} through the Groq API, with temperature set to $0.7$ and a maximum generation length of 300 tokens. Rate limiting and retry handling are applied during generation.

After augmentation, the original 2,764 training instances expand to 8,292 instances. A cleaning step removes malformed or low-quality outputs, yielding 8,282 final training instances. The original set of 137 users is preserved, and no new timestamps or emotion trajectories are created. Thus, the augmentation increases lexical variation but does not add new temporal supervision.

\subsection{Interpretation}
\label{app:augmentation_interpretation}

The augmentation and transfer learning experiments are interpreted as auxiliary analyses. EmoBank transfer tests whether external dimensional affect supervision improves representation learning, while LLM-based augmentation tests whether paraphrased lexical variation improves prediction. Neither strategy changes the longitudinal structure of the dataset. Consequently, these experiments cannot directly address the main temporal modeling challenge, which requires user-specific affective histories rather than additional isolated text examples.

\section{Forecasting Variant Details}
\label{app:forecasting_variants}

This appendix provides additional implementation details for the affect forecasting variants described in Section~\ref{subsubsec:forecast_models}. These variants are used to isolate the effects of text, user information, temporal feature design, loss function, and dimension-specific modeling in next-step affective change forecasting.

\subsection{Forecasting target construction}
\label{app:forecasting_target_construction}

For each user $u$, observations are sorted chronologically. The forecasting target is the signed change between consecutive affective states:
\[
\Delta^{(v)}_{u,i}=v_{u,i+1}-v_{u,i},
\qquad
\Delta^{(a)}_{u,i}=a_{u,i+1}-a_{u,i}.
\]
The final observation for each user has no future observation and is therefore excluded from supervised training. This construction produces 2,627 labeled transition rows from the 2,764 training observations. For model selection, the last labeled transition for each user is reserved for validation, yielding one validation target per user.

\subsection{Base temporal features}
\label{app:base_temporal_features}

The base temporal feature vector contains compact numeric information about the current and immediately preceding affective trajectory. It includes the current valence and arousal values, previous valence and arousal changes, absolute previous changes as volatility indicators, and a cold-start indicator for the beginning of a user trajectory. Boundary values with no available history are imputed with zero.

All temporal features are standardized using a \texttt{StandardScaler} fitted only on the inner training split. The same scaler is then applied to validation and test features.

\subsection{ACF-Numeric+ objective}
\label{app:acf_numeric_plus}

ACF-Numeric+ is an auxiliary numeric forecasting variant that modifies the standard numeric model by replacing the mean squared error objective with a hybrid SmoothL1 and Pearson-correlation objective:
\[
\mathcal{L}_{\text{hybrid}}
=
0.7\,\mathcal{L}_{\text{SmoothL1}}
+
0.3\,(1-r),
\]
where $r$ denotes the Pearson correlation between predicted and true affective changes within a mini-batch. The SmoothL1 term encourages low absolute error, while the correlation term encourages stronger rank-order agreement between predicted and observed affective changes.

This variant is included to test whether optimizing for both magnitude accuracy and correlation improves forecasting. It is treated as an auxiliary ablation variant rather than the final forecasting model.

\subsection{ACF-Residual}
\label{app:acf_residual_details}

ACF-Residual decomposes valence forecasting into a linear autoregressive component and a nonlinear residual correction. The linear component is fitted using scikit-learn Ridge regression with default regularization strength $\alpha=1.0$. The residual corrector is then trained to model the remaining error:
\[
\hat{\Delta}^{(v)}_{\text{residual}}
=
\Delta^{(v)}
-
\hat{\Delta}^{(v)}_{\text{linear}}.
\]
The final valence prediction is the sum of the Ridge prediction and the neural residual prediction:
\[
\hat{\Delta}^{(v)}_{\text{ACF-Residual}}
=
\hat{\Delta}^{(v)}_{\text{linear}}
+
\hat{\Delta}^{(v)}_{\text{residual}}.
\]

A mild post-hoc variance calibration is applied to valence predictions:
\[
\hat{\Delta}^{(v)}_{\text{calibrated}}
=
\alpha_v
\hat{\Delta}^{(v)}_{\text{ACF-Residual}},
\qquad
\alpha_v=1.15.
\]
This calibration expands prediction variance and reduces valence MAE without changing Pearson correlation.

\subsection{ACF-Temporal}
\label{app:acf_temporal_details}

ACF-Temporal extends the base temporal feature representation with richer trajectory descriptors. These include lagged affective changes, second-order change terms, acceleration features, rolling statistics over recent observations, and direction-consistency indicators. These features are designed to capture short-term momentum, volatility, and local trajectory shape.

ACF-Temporal is used as the final arousal forecaster because arousal benefits more from the richer temporal representation than from the residual decomposition used for valence.

\subsection{ACF-Hybrid}
\label{app:acf_hybrid_details}

ACF-Hybrid combines the strongest dimension-specific forecasting variants:
\[
\hat{\Delta}_{\text{Hybrid}}
=
\left(
\hat{\Delta}^{(v)}_{\text{ACF-Residual}},
\hat{\Delta}^{(a)}_{\text{ACF-Temporal}}
\right).
\]
The final hybrid model therefore uses ACF-Residual for valence and ACF-Temporal for arousal. It does not use text representations or learned user embeddings in the final forecasting framework. Instead, it relies on numeric affective history and dimension-specific temporal design.

\end{document}